\documentclass[10pt,twocolumn,letterpaper]{article}

\usepackage{cvpr}              
\usepackage{booktabs}
\usepackage{multirow}
\usepackage{array}
\usepackage{listings}
\usepackage{xcolor}
\usepackage{graphicx}
\usepackage[most]{tcolorbox}
\usepackage{pgf-pie}
\usepackage{tikz}
\usetikzlibrary{arrows.meta}
\usetikzlibrary{shapes.geometric, positioning}
\usepackage{subcaption}
\usepackage{float}
\usepackage{adjustbox}
\usepackage{graphicx}

\newtcolorbox{conversationbox}[2][]{%
    colback=white,
    colframe=black!70,
    boxrule=1pt,
    title=\textbf{#2},
    left=1mm, 
    right=1mm,
    top=1mm, 
    bottom=1mm,
    before skip=5pt,
    after skip=0pt,
    breakable=false,
    sharp corners=south, 
    #1
}

\newtcolorbox{subconversationbox}[1][]{%
    colback=white,
    colframe=black!70,
    boxrule=1pt,
    left=1mm, 
    right=1mm,
    top=1mm, 
    bottom=1mm,
    before skip=0pt, 
    after skip=5pt,
    breakable=false,
    sharp corners=north,
    #1
}

\definecolor{cvprblue}{rgb}{0.21,0.49,0.74}
\usepackage[pagebackref,breaklinks,colorlinks,allcolors=cvprblue]{hyperref}

\def\paperID{6645} 
\def\confName{CVPR}
\def\confYear{2025}

\title{Maya: An Instruction Finetuned Multilingual Multimodal Model}

\author{
Nahid Alam$^{1*,2}$, 
\and 
Karthik Reddy Kanjula$^2$,
\and 
Surya Guthikonda$^{3,2}$,
\and 
Timothy Chung$^{4,2}$,
\and 
Bala Krishna S Vegesna$^5$,
\and 
Abhipsha Das$^2$,
\and 
Anthony Susevski$^2$,
\and 
Ryan Sze-Yin Chan$^6$,
\and 
S M Iftekhar Uddin$^2$,
\and 
Shayekh Bin Islam$^{7,2}$,
\and 
Roshan Santhosh$^8$,
\and 
Snegha A$^{9}$,
\and 
Drishti Sharma$^2$,
\and 
Chen Liu$^{10}$,
\and 
Isha Chaturvedi$^{11*}$,
\and 
Genta Indra Winata$^{12*}$,
\and 
Ashvanth.S$^2$,
\and 
Snehanshu Mukherjee$^{13}$,
\and 
Alham Fikri Aji$^{14}$
\and 
\\
$^1$Cisco Meraki, 
$^2$Cohere For AI Community, 
$^3$Indiana University Bloomington, \\
$^4$Imperial College London,
$^5$Georgia Institute of Technology,
$^6$The Alan Turing Institute, \\
$^7$Bangladesh University of Engineering and Technology, 
$^8$University of Pennsylvania, 
$^9$IIT Bombay, \\
$^{10}$TU Darmstadt,
$^{11}$Articul8 AI, 
$^{12}$Capital One,
$^{13}$IIT Dhanbad, 
$^{14}$MBZUAI \\
\\
{\tt\small {nahid.m.alam@gmail.com}}
}

\begin{document}
\maketitle
\renewcommand\thefootnote{\fnsymbol{footnote}} 
\footnotetext[1]{Work does not relate to the authors' positions.}
\begin{abstract}
The rapid development of large Vision-Language Models (VLMs) has led to impressive results on academic benchmarks, primarily in widely spoken languages. However, significant gaps remain in the ability of current VLMs to handle low-resource languages and varied cultural contexts, largely due to a lack of high-quality, diverse, and safety-vetted data. Consequently, these models often struggle to understand low-resource languages and cultural nuances in a manner free from toxicity. To address these limitations, we introduce Maya, an open-source Multimodal Multilingual model. Our contributions are threefold: 1) a multilingual image-text pretraining dataset in eight languages, based on the LLaVA pretraining dataset; 2) a thorough analysis of toxicity within the LLaVA dataset, followed by the creation of a novel toxicity-free version across eight languages; and 3) a multilingual image-text model supporting these languages, enhancing cultural and linguistic comprehension in vision-language tasks. Code available at \url{https://github.com/nahidalam/maya}.
\end{abstract}
\section{Introduction}
\label{sec:intro}

Vision Language Models (VLMs) have emerged as a key technique in artificial intelligence, enabling machines to understand and reason about the visual world through natural language. Recent advancements in Large Language Models (LLMs) and general-purpose image encoders such as CLIP \citep{radford2021learning} and SigLIP \citep{zhai2023sigmoid} has significantly boosted VLM capabilities. Architectures such as Flamingo \citep{alayrac2022flamingo}, LLaVA \citep{liu2023llava, liu2023improvedllava}, KOSMOS \citep{kosmos-g, peng2023kosmos}, Florence-2 \citep{xiao2024florence} and Molmo \citep{deitke2024molmo} demonstrate strong performance across tasks including image captioning, Visual Question Answering (VQA), and complex reasoning. Qwen2-VL \citep{wang2024qwen2} introduced Multimodal Rotary Position Embedding (M-RoPE) \citep{su2021roformer} and dynamic resolution techniques, while PaLI’s joint modality scaling and cross-lingual learning \citep{chen2022pali, chen2023pali} have furthered vision-language understanding. Despite these advances, VLMs remain limited to high-resource languages. This creates a disparity in accessibility, as current models struggle with cultural contexts and visual concepts in low-resource languages \citep{joshi2020state}. 

This gap is largely due to the lack of high-quality multilingual multimodal datasets. Pre-training datasets like COCO \citep{lin2014microsoft}, Flickr30K \citep{young2014image}, LAION \citep{schuhmann2022laion}, Visual Genome \citep{krishna2017visual}, and LLaVA \citep{liu2023llava} are predominantly in English, limiting cross-linguistic generalization. Existing multilingual datasets, such as Multi30k \citep{elliott2016multi30k} and Crossmodal-3600 \citep{thapliyal2022crossmodal}, are limited in scale and cultural diversity. Current datasets often contain toxic and culturally insensitive content \cite{yue2024pangea}, perpetuating biases and stereotypes. To our knowledge, no peer-reviewed research has systematically addressed the mitigation of toxicity and harmful content in image-text datasets. This highlights the need for comprehensive, toxicity-mitigated datasets that support robust, multilingual training while capturing linguistic and cultural diversity.

To address these challenges, we introduce Maya, an open-source Multilingual Multimodal Vision Language Model (mVLM) that expands multimodal capabilities to eight languages with an emphasis on data quality and cultural sensitivity. Built on the LLaVA framework, Maya includes a newly created pre-training dataset designed to support multilingual and culturally aware VLM development. Our key contributions include: 
\begin{enumerate}
    \item a novel multilingual image-text pretraining dataset consisting of \emph{558,000} images for future development of mVLMs, 
    \item a toxicity-free version of the dataset in 8 languages, and 
    \item a new mVLM that demonstrates improved performance in understanding cultural and linguistic nuances compared to PALO-7B \cite{maaz2024palo} on LLaVA-Bench-In-The-Wild \cite{liu2023llava}, offering a multilingual alternative to LLaVA \cite{liu2023llava}.
\end{enumerate}


\section{Related Work}
\label{sec:relatedwork}

\subsection{Multilingual Large Language Models}

While open-source large language models have made notable strides, progress in multilingual LLMs (mLLMs) still lags. Many LLMs \cite{team2024gemma2, dubey2024llama, abdin2024phi, ministral} offer limited multilingual support, typically focusing on high-resource languages with reduced performance in others. Dedicated mLLMs \cite{xue2020mt5, lin2021few, le2023bloom, qwen, lin2024mala, ustun2024aya, aryabumi2024aya}, such as BLOOM \citep{le2023bloom} and Aya-101 \citep{ustun2024aya}, extend language coverage to 43 and 101 languages respectively, improving performance across tasks like translation, summarization, and reasoning.

\subsection{Multimodal Large Language Models}

Research in Multimodal Large Language Models (MLLMs) has advanced significantly in enabling models to process both images and text across diverse tasks, with progress along two main paths: proprietary closed-source systems \cite{gpt4o, team2023gemini, claude} and open-source initiatives. While most of the initial efforts focused predominantly on English-language capabilities, such as LLaVA \cite {liu2023llava, liu2023improvedllava}, several recent works have attempted to close this gap by targeting non-English languages. PALO \cite{maaz2024palo} extended LLaVA \cite{liu2023llava} to cover 10 major languages via semi-automated translations of the LLaVA-Instruct-150k dataset \cite {liu2023llava}. X-LLaVA \cite{shin2024x} demonstrated strong Korean-English bilingual performance. Broader language coverage has been achieved by Llama 3.2 \cite{dubey2024llama} supporting 8 languages and Qwen2-VL \cite{wang2024qwen2} covering several European languages alongside Japanese, Korean, Arabic and Vietnamese. More recent developments have pushed multilingual boundaries through diverse approaches: PaliGemma \cite{beyer2024paligemma} pretrained on a variety of vision-language tasks upto 35 languages providing an open base MLLM but without instruction tuning, \cite{geigle2023mblip} mBLIP realigned English-tuned image encoder using machine-translated data for 95 languages, Pangea \cite{yue2024pangea} is a multilingual multimodal model trained on an instruction dataset spanning 39 languages, Parrot \cite{sun2024parrotmultilingualvisualinstruction} addressed supervised fine-tuning imbalances across 6 languages through textual guidance for visual token alignment, while models like Phi-3.5-Vision \cite{abdin2024phi}, Pixtral \cite{agrawal2024pixtral} and Molmo \cite{deitke2024molmo} are mostly English-centric \cite{yue2024pangea}.

\subsection{Multimodal Multilingual Datasets}

Large-scale datasets for retrieval and captioning such as xFlickrCO \citep{bugliarello-etal-2022-iglue}, BLIP3-KALE \citep{awadalla2024blip3kaleknowledgeaugmentedlargescale} and WIT \citep{wit_2021} have been developed to support broader multimodal research. For instruction tuning, PALO \cite{maaz2024palo} built a dataset of 150K images in 10 languages from a subset of LLaVA image-text instruction tuning dataset on English text, while Pangea \cite{yue2024pangea} demonstrated broader coverage with image-text pair dataset in 39 languages.

\subsection{Evaluating Multimodal Models}

Evaluating multilingual multimodal LLMs \cite{beyer2024paligemma, wang2024qwen2, yue2024pangea} involves two main categories. The first assesses standard multimodal tasks, such as cross-lingual visual question answering (VQA) with xGQA \citep{pfeiffer2021xgqa} and image-text retrieval via xFlickrCO \citep{bugliarello-etal-2022-iglue}. The second evaluates cultural understanding, using datasets like MaRVL \citep{liu-etal-2021-visually} for visual reasoning, CVQA \citep{romero2024cvqa} for cultural context, and XM3600 \citep{thapliyal2022crossmodal} for cross-modal comprehension. Significant progress has been made in developing multilingual multimodal datasets for evaluating VLMs including XVNLI \citep{bugliarello-etal-2022-iglue} for visual natural language inference, MTVQA \citep{tang2024mtvqa} for temporal reasoning, M3Exam \citep{zhang2023m3exam} for educational assessment, PangeaBench \cite{yue2024pangea} for broad multilingual capabilities, MMMB \cite{sun2024parrotmultilingualvisualinstruction} for cross-lingual understanding, and LLaVA-Bench-In-The-Wild \cite{liu2023llava, maaz2024palo} for real-world scenarios. Notably, none of these datasets focus on evaluating toxicity.


\section{Dataset Creation and Filtering}
\label{sec:dataset}
\begin{figure*}[t]
  \centering
  \fbox{\includegraphics[width=0.95\textwidth]{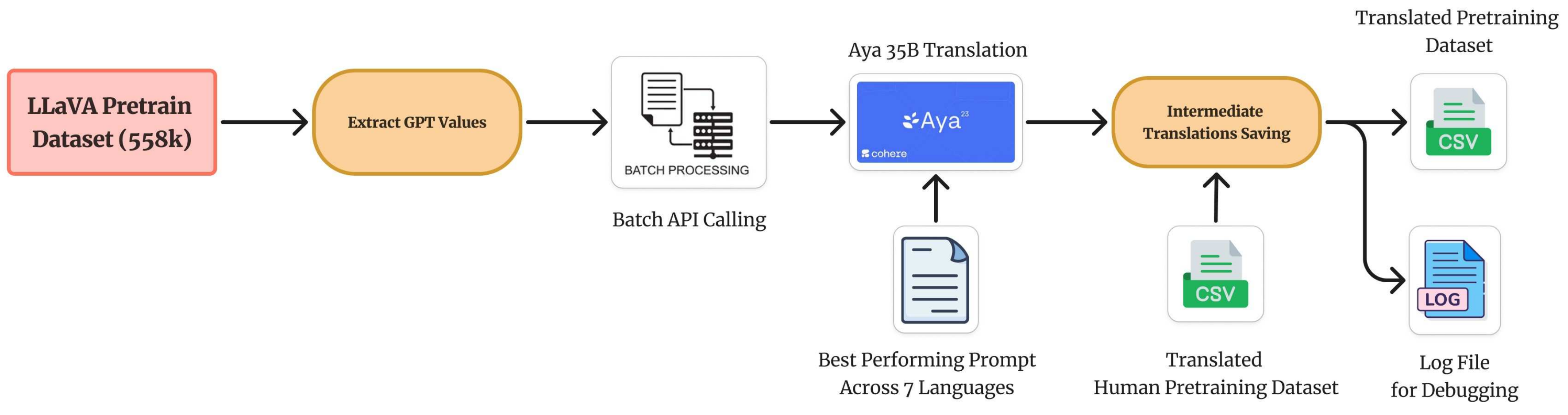}}
  \caption{Pretrain Dataset Preparation Process}
  \label{fig:createdata}
\end{figure*}
\subsection{Dataset Creation Methodology}

 Recently, PALO \cite{maaz2024palo}  and Pangea \cite{yue2024pangea} have created multilingual image-text dataset for building multilingual multimodal models. However, these multilingual datasets often suffer from data quality issues and distribution biases across languages. For instance, in the PALO dataset, the distribution varies significantly between English and other languages \cite{elliott2016multi30k, hinck2024llava, maaz2024palo}. To address these limitations, we present a novel pre-training dataset tailored for LLaVA's architecture that is both multilingual and optimized for diverse language representation. Our dataset introduces rigorous processes for toxicity analysis, distribution balancing, and quality control, ensuring consistent and reliable performance across multiple languages and modalities as shown in Figure \ref{fig:createdata}. We expand the original English LLaVA dataset, which contains 558,000 samples, to include seven additional languages—Chinese, French, Spanish, Russian, Hindi, Japanese, and Arabic—yielding a total of 4.4 million samples, equally distributed across all eight languages. Every sample in our dataset has undergone toxicity verification and filtering, ensuring a safer training corpus than the original LLaVA pretraining dataset.

Our approach encompasses three core components: 1) parallel dataset creation using a hybrid translation method, 2) prompt engineering optimization, and 3) scalable generation of pre-training datasets. This pipeline integrates multiple language models, such as \cite{gpt4o, team2023gemini, claude}, alongside specialized multilingual models like Aya 35B \cite{aryabumi2024aya}, to ensure high-quality, cross-lingual data suitable for multilingual applications. 

\subsubsection{Initial Dataset Processing}
Our pipeline is built on the LLaVA dataset \cite{liu2023llava}, centered around image-text pairs and their corresponding GPT responses. We use stratified sampling to select 30 representative samples per language, optimizing for linguistic diversity through Length Analysis (LA), Flesch Reading Ease (FRE), and Flesch-Kincaid Grade Level (FKGL) metrics \citep{textstat}. To ensure quality, we implement a cascaded translation verification system. First, we use Google Translate to do initial translation. This is followed by back-translation verification. Finally human reviewers verify it with the help of \citep{claude, team2023gemini, achiam2023gpt}.

\subsubsection{Prompt Engineering and Evaluation}

\begin{figure}[t]
  \centering
  \fbox{\adjustbox{trim=0pt 0pt 0pt 0pt, clip}{\includegraphics[width=1\linewidth]{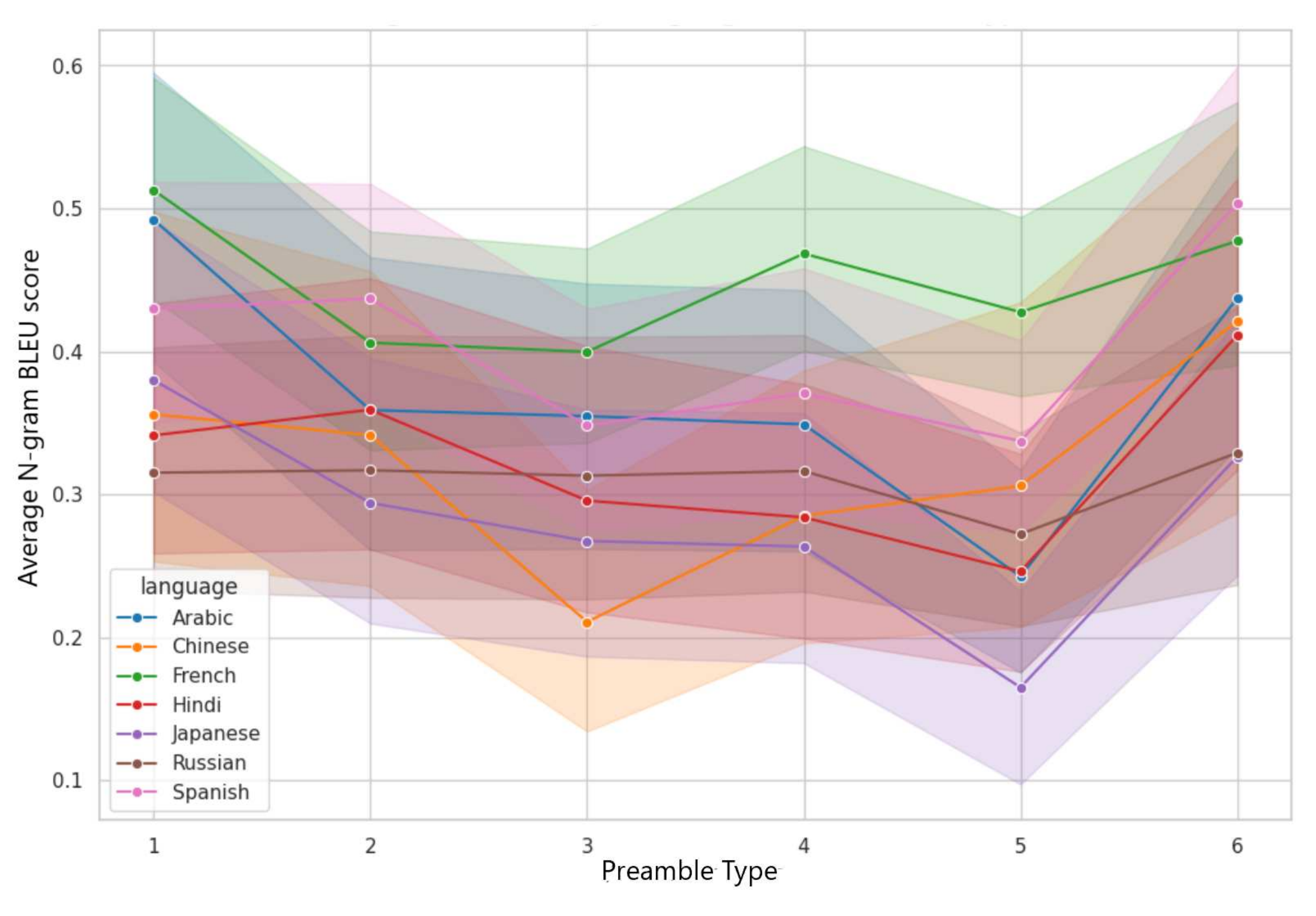}}}
  \caption{N-gram values by language and preamble type. These values are average of 1-gram, 2-gram, 3-gram and 4-gram}
  \label{fig:preamble_eval}
\end{figure}

\begin{figure}[t]
  \centering
  \fbox{\includegraphics[width=0.9\linewidth]{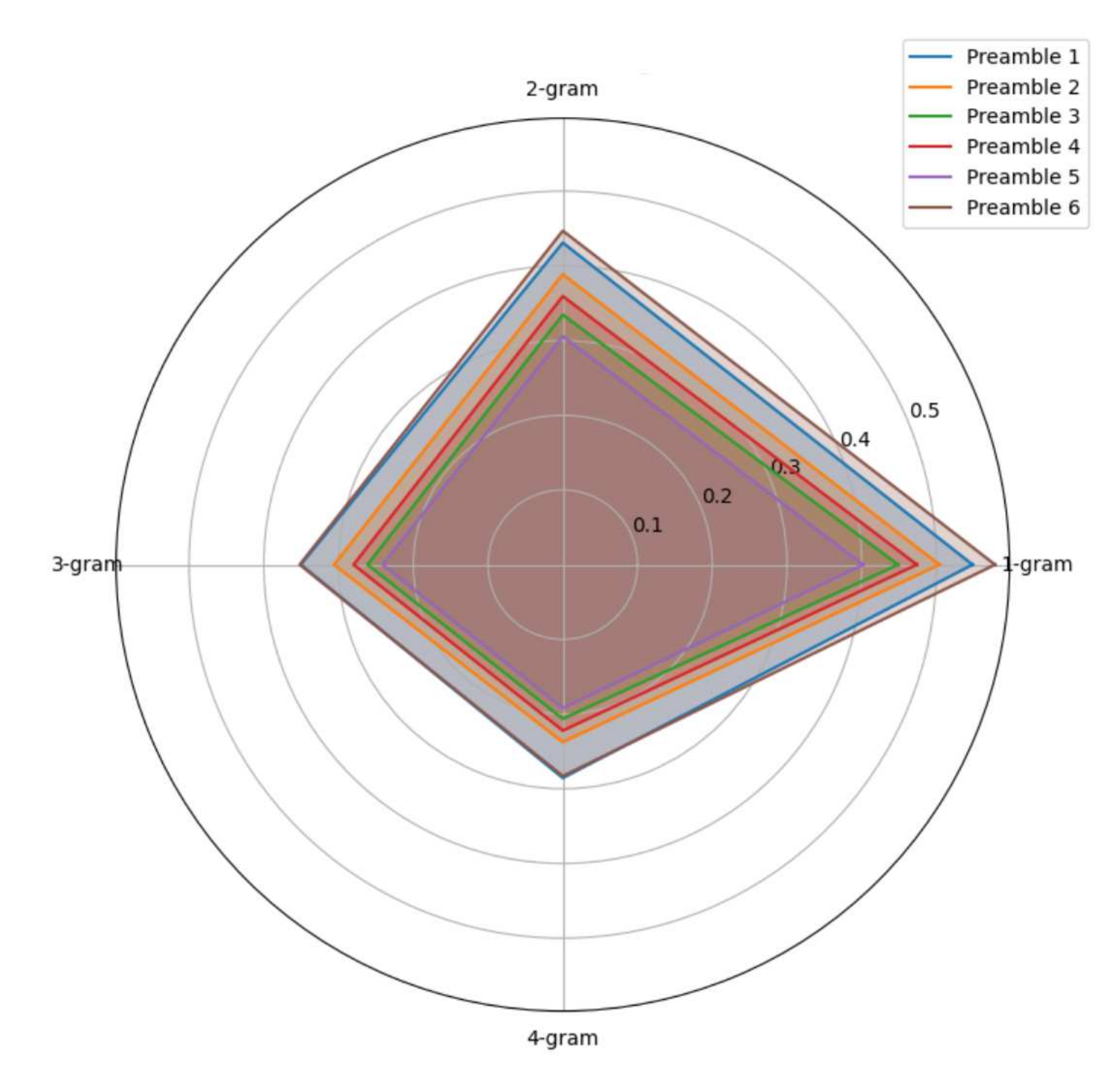}}
  \caption{Radar chart of N-gram averages by preamble}
  \label{fig:preamble_radar}
\end{figure}

In the prompt engineering phase, we employ a prompt evaluation process for each language. First, we create a prompt evaluation dataset using a similar process as Figure \ref{fig:createdata}. We take 6 sample prompts and translate the English text of the prompt evaluation dataset to seven languages using Aya 35B.  We then compare these translations with the reference translation of the prompt evaluation dataset using BLEU score \cite{papineni2002bleu} and N-gram score \cite{shannon1948mathematical, brill-etal-1998-beyond}. In Figure \ref{fig:preamble_eval}, we plot the n-gram BLEU scores (averaged across 1-gram, 2-gram, 3-gram, and 4-gram) for each language and preamble type.  We find that across all seven languages (Arabic, Chinese, French, Hindi, Japanese, Russian, and Spanish), Preamble 6 consistently achieves the highest N-gram BLEU scores, with most languages showing a notable upward trend from Type 5 to Type 6, reaching values around 0.4-0.5. The radar plot in Figure \ref{fig:preamble_radar} provides complementary evidence by breaking down the N-gram analysis (1-gram through 4-gram) averaged across all languages for each preamble type. It reveals that Preamble 6 maintains the largest area coverage, indicating superior performance across all N-gram levels compared to other preamble types. This is particularly significant as higher N-gram BLEU scores suggest better preservation of phrase structures and contextual meaning in the translations. The relatively balanced performance of Preamble 6 across different languages  and across different N-gram levels indicates its robustness and language-agnostic effectiveness in the translation task. We find that preamble 6 is the best preamble across all the seven languages. The resulting prompt template is in listing~\ref{lst:translation_instructions}. We use this prompt template in our translation framework. 
\lstset{
    basicstyle=\ttfamily\footnotesize,       
    backgroundcolor=\color{black!5},  
    frame=single,                     
    keywordstyle=\bfseries,           
    morekeywords={Input, Expected, Ensure, Note, Instructions, Examples},
    breaklines=true,                  
    breakatwhitespace=false,          
    columns=fullflexible              
}

\begin{figure}[t]
\begin{lstlisting}[caption={Translation Instructions}, label={lst:translation_instructions}]
## Instructions
You are an expert in translations. 
Your job is to translate the input to Japanese in
the given chat.

Ensure that:
{Specific Things to Consider while Translating}

Note: {Extra Constraints on Output Generation}

## Examples
### Example 1
Input:
{Input Sentence}
Expected Output:
{Translated Sentence}
\end{lstlisting}
\end{figure}

\subsubsection{Translation Framework Design}

Our translation framework uses the best preamble identified in the prompt engineering evaluation step. The prompt includes 1) standardized input-output formatting to maintain uniformity across languages, 2) example-driven instruction sets tailored for complex translations, and 3) integrated validation triggers to ensure quality control. Leveraging Aya 35B advanced translation capabilities, our framework achieves greater than 0.47 average BLEU scores across seven languages.

\subsubsection{Scalable Dataset Generation}

For large-scale dataset generation, we implement a batch processing pipeline integrated with Aya 35B API as shown in Figure \ref{fig:createdata}. First, we extract GPT values with quality filter from the LLaVA pretrain dataset. These extracted GPT values are then passed through the Aya 35B batch-optimized API calling with intermediate translation checkpointing. The pipeline does necessary error handling and comprehensive logging for quality tracking. This architecture enables efficient processing of the full 558K samples while maintaining translation quality. We implement version control for intermediate translations and maintain detailed debug logs, ensuring reproducibility and enabling systematic error analysis.

\subsection{Dataset Toxicity Filtering}
In the original English LLaVA Pretrain dataset which has 558,000 image-caption pairs, we observed some toxic contents. To methodically identify toxic content, we used LLaVAGuard 7B \cite{helff2024llavaguard} and Toxic-BERT \cite{Detoxify}. For images, we relied on the LlavaGuard 7B framework to spot and categorize unsafe or toxic visuals based on set guidelines. For text, we used the Toxic-BERT model to scan captions and flag anything with offensive or harmful language.

\begin{figure}[t]
  \centering
  \fbox{\includegraphics[width=1\linewidth]{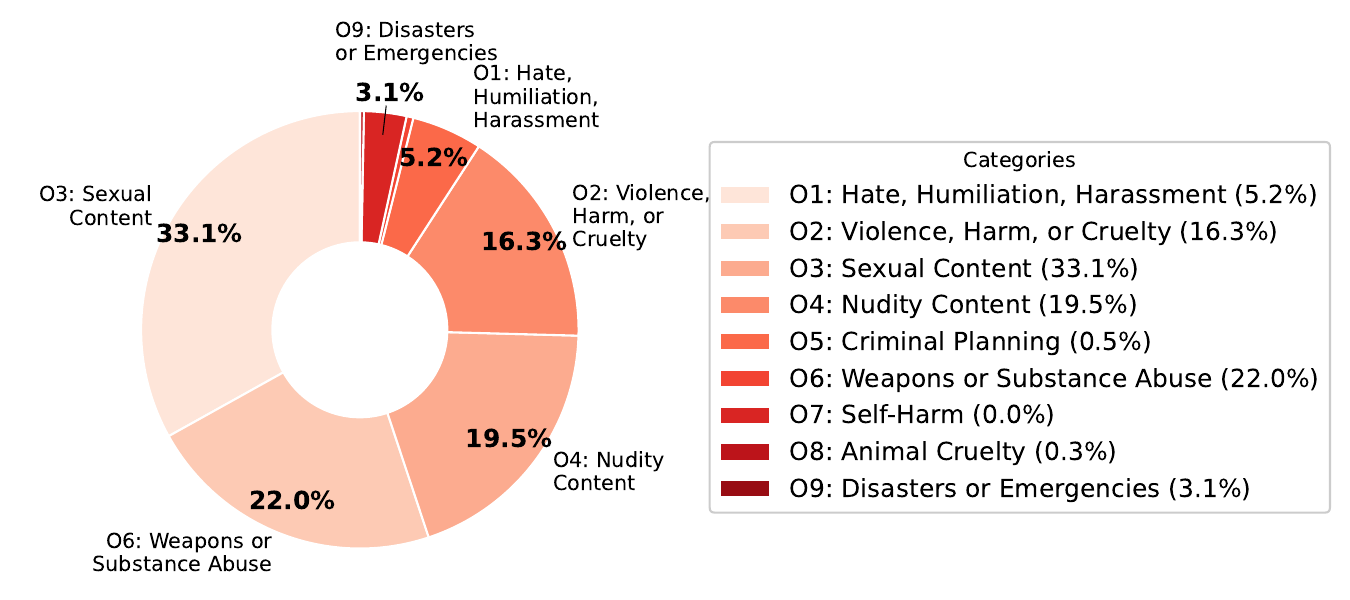}}
  \caption{Image Toxicity Analysis with LLaVAGuard}
  \label{fig:toxanalysis}
\end{figure}

LLaVAGuard is specifically designed for in-depth toxicity analysis, assigning safety ratings, categories, and rationales to flagged images, producing structured data about each identified risk. This analysis led to the categorization of content across various types, including sexual content, hate speech, violence, and substance abuse, which are visually represented in the toxicity analysis chart as shown in Figure \ref{fig:toxanalysis}. Additionally, a prompt optimization pipeline was introduced to ensure precise filtering, capturing potentially harmful content while reducing false positives.

\begin{figure}[t]
  \centering
  \fbox{\includegraphics[width=1\linewidth]{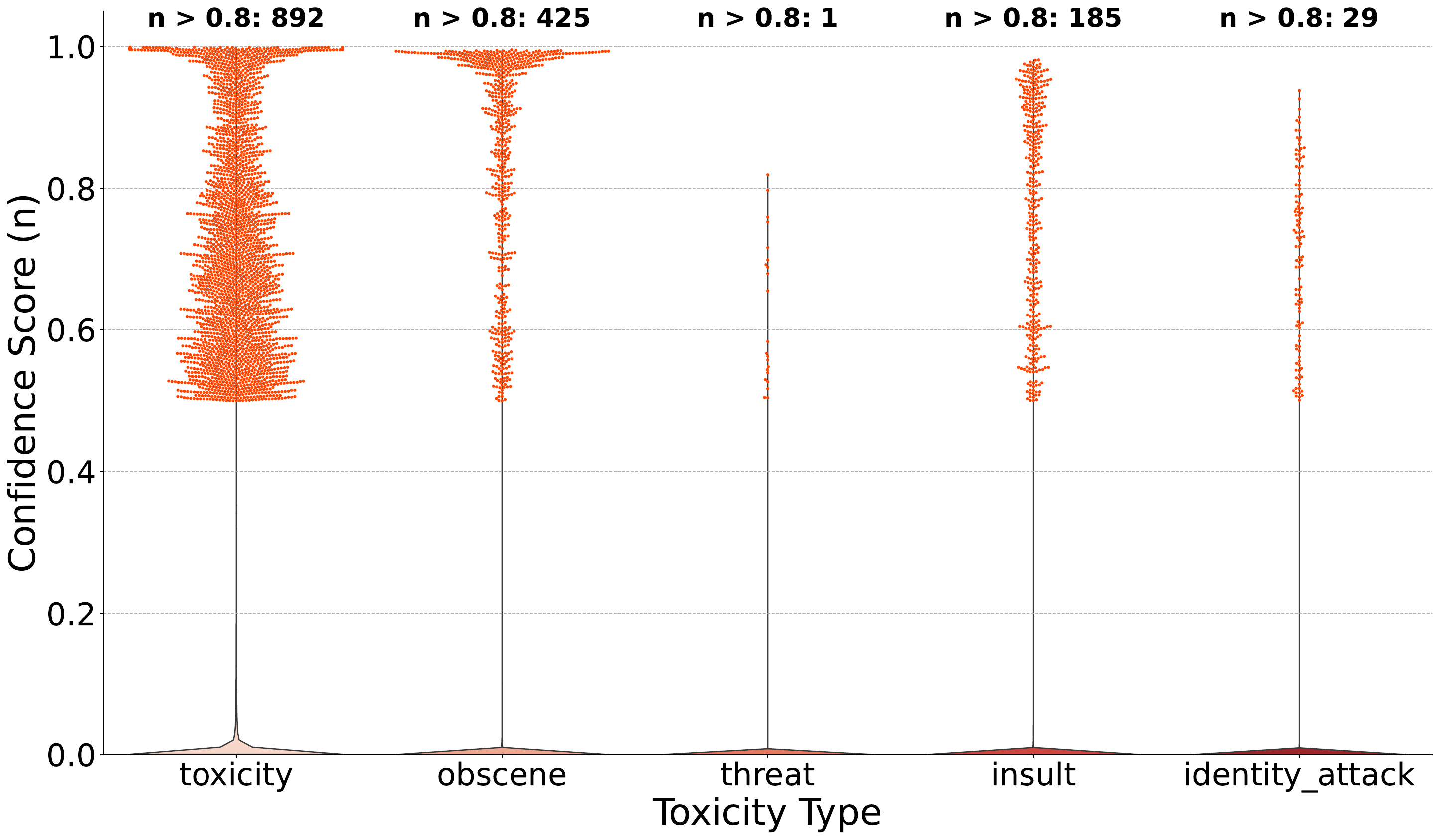}}
  \caption{Image Caption Toxicity Analysis with Toxic-BERT}
  \label{fig:toxbert}
\end{figure}

\begin{figure}[t]
  \centering
  \fbox{\includegraphics[width=\linewidth]{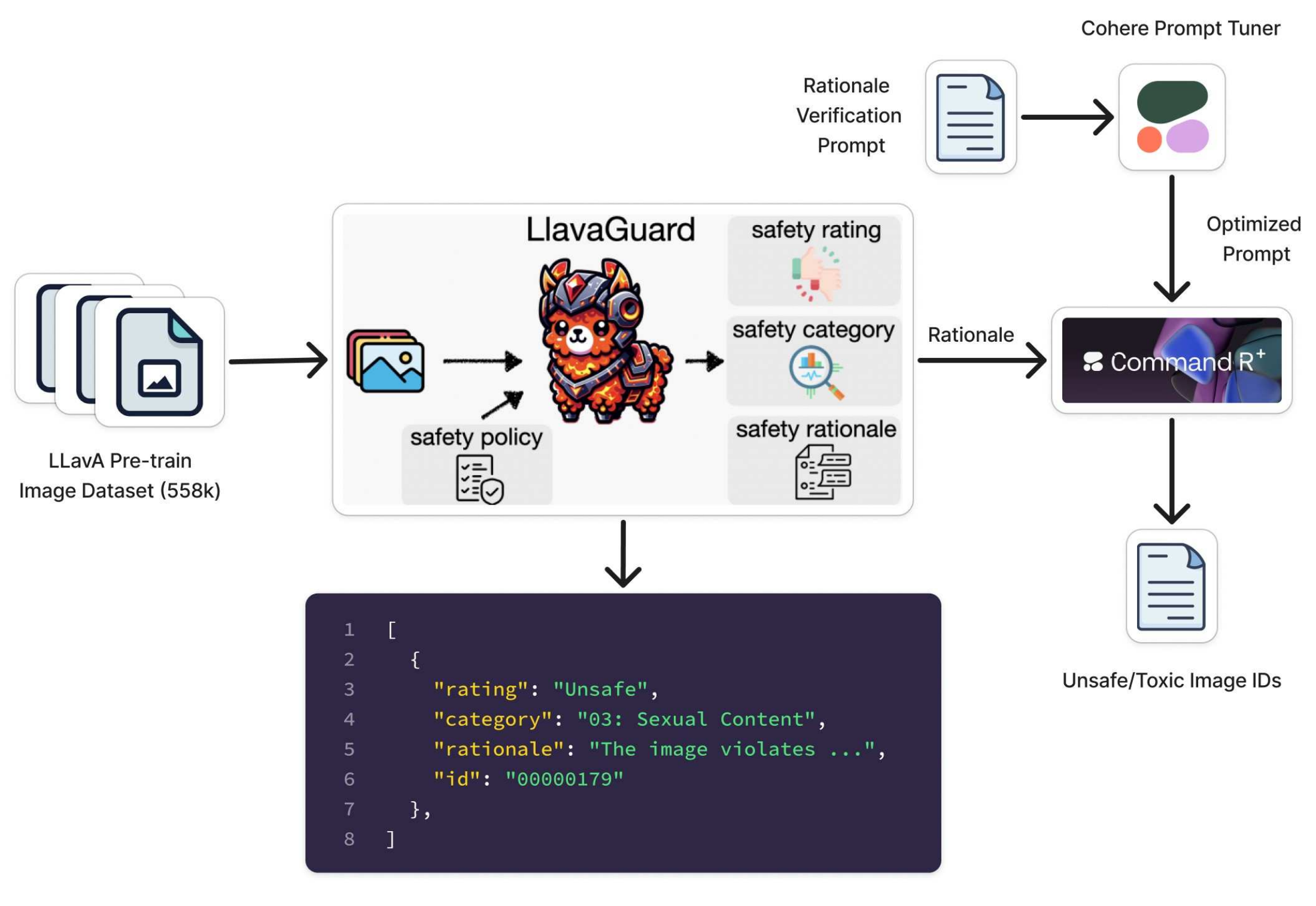}}
  \caption{Dataset Toxicity Filtering Method}
  \label{fig:toxfilter}
\end{figure}

The overall process for creating toxicity free dataset is shown in Figure \ref{fig:toxfilter}. LLaVaGuard output provides a rating, category, and rationale explaining why an image violates guidelines. We then refine this by identifying genuinely toxic images. This is done by developing an optimized prompt using the Cohere prompt tuner\footnote{\url{https://docs.cohere.com/v2/docs/prompt-tuner}}. We use this prompt as a preamble (System Prompt) and pair it with the previous information—rating, category, and rationale—associated with each image flagged as unsafe by LLaVAGuard. Command R+ \cite{command_r} then analyzes these results to identify the truly unsafe image IDs. In our analysis, LLaVAGuard identified 7,600 images as toxic and the final Command R+ output identified  7,111 images as unsafe. 

For identifying toxic captions, we used Toxic-BERT that categorized 892 image captions as toxic with greater than 80\% confidence as shown in Figure \ref{fig:toxbert}. LLaVAGuard and Command R+ identified 7,111 images and Toxic-BERT identified 892 images, there are in total 7,531 unique toxic images. We then removed those 7,531 images from the pretraining dataset to create the toxicity free pretraining dataset.

\section{Multilingual Multimodal Instruction Tuning}
\label{sec:model}

\subsection{Model Architecture}

The Maya model architecture draws inspiration from LLaVA 1.5 \cite{liu2023improvedllava}. Our goal is to leverage pretrained multilingual Aya model \cite{aryabumi2024aya} and vision encoder with multilingual capabilities. Specifically, we employ the Aya-23 8B model as our LLM $f_{\phi}$ because of its multilingual capability. Aya-23 has 8 billion parameters, an 8K context window, and is trained across 23 languages. Our dataset, however, includes 8 of these 23 languages, aligning with our objective of optimizing Maya for a diverse yet focused linguistic range.

For the vision encoder, we opted for SigLIP\footnote{\texttt{siglip-base-patch16-256-multilingual} from \url{https://huggingface.co/google/siglip-base-patch16-256-multilingual}} \cite{zhai2023sigmoid} rather than CLIP \cite{radford2021learning}, which is traditionally used in LLaVA. This choice is motivated by SigLIP's strong performance, multilingual adaptability, and capacity for variable-length patch sizes. Unlike CLIP, SigLIP supports scalable positional embeddings, enabling it to accept inputs of varying dimensions through positional embedding interpolation. This flexibility makes SigLIP particularly suitable for our purpose. For each input image $X_v$, we get the visual features from SigLIP, $Z_v = g(X_v)$. A trainable projection matrix $W$ then converts the image features $Z_v$ into language features $H_v$. 

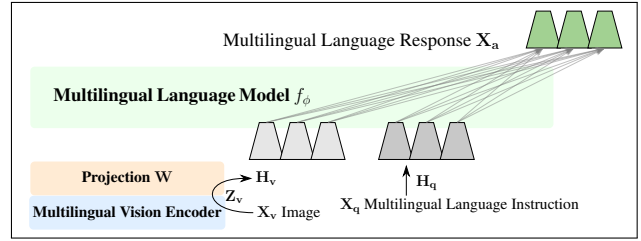
\begin{figure}[t]
    \centering
    \fbox{
    \resizebox{0.95\linewidth}{!}{ 
    \begin{tikzpicture}[
      node distance=1.4cm and 1.4cm,
      every node/.style={align=center},
      text height=2ex, text depth=.35ex,
      arrow style/.style={-{Stealth[length=3mm, width=2mm]}, thick, draw=black!50!white, opacity=0.5}
    ]
    
    \definecolor{lightgreen}{RGB}{236, 255, 236}
    \definecolor{lightblue}{RGB}{224, 239, 255}
    \definecolor{lightorange}{RGB}{255, 235, 212}
    \definecolor{greenbox}{RGB}{162, 209, 146}
    \definecolor{lightgrey1}{RGB}{230, 230, 230}
    \definecolor{lightgrey2}{RGB}{200, 200, 200}
    
    \node[fill=lightgreen, rounded corners, minimum width=17cm, minimum height=2cm, anchor=north west, align=left, text width=15.5cm] 
        (lm-bg) at (-6.2,0) {\LARGE \textbf{Multilingual Language Model }\( f_{\phi} \)};
    
    \node[fill=lightorange, rounded corners, minimum width=6.3cm, minimum height=1.1cm, anchor=south west] 
    (projection) at (-6.2,-4.2) {\Large\textbf{Projection} \( \mathbf{W} \)};
    \node[fill=lightblue, rounded corners, minimum width=6.3cm, minimum height=1.1cm, below=0cm of projection, align=left] 
    (vision-encoder) {\Large\textbf{Multilingual Vision Encoder}};
    
    \node[right=0.9cm of projection, anchor=west] (hv) {\Large \( \mathbf{H_v} \)};
    \node[right=0.9cm of vision-encoder, anchor=west] (xv) {\Large \( \mathbf{X_v} \) Image};
    
    \draw[-{Stealth[length=4mm, width=2.5mm]}, thick]
        (xv.west) 
        .. controls ++(-1.7,0.0) and ++(-1.7,0.0) 
        .. (hv.west);
    
    \node[above left= -0.1cm and 0.2cm of xv, align=center] (zv) {\Large \( \mathbf{Z_v} \)};
    
    \foreach \i in {0,1,2} {
        \node[fill=lightgrey1, draw=black, trapezium, trapezium angle=75, minimum width=1cm, minimum height=1.2cm, above=0.3cm of hv, xshift=\i cm] (hexagon\i) {};
    }
    
    \foreach \i in {3,4,5} {
        \node[fill=lightgrey2, draw=black, trapezium, trapezium angle=75, minimum width=1cm, minimum height=1.2cm, right=1.4cm of hexagon2, xshift=(\i-3)*1cm] (hexagon\i) {};
    }
    
    \node[below=0.4cm of hexagon4, anchor=north] (hq) {\Large \( \mathbf{H_q} \)};
    \node[below right=1.23cm and 1cm of hexagon3, anchor=north] (xq) {\quad\qquad\Large \( \mathbf{X_q} \) Multilingual Language Instruction};
    
    \draw[-{Stealth[length=4mm, width=2.5mm]}, thick] ([xshift=-1.2cm]xq.north) -- ++(0,1);
    
    \node[above left=2.6cm and 1.8cm of hexagon4, anchor=south] (xa) {\LARGE Multilingual Language Response \( \mathbf{X_a} \)};
    
    \foreach \i in {6,7,8} {
        \node[fill=greenbox, draw=black, trapezium, trapezium angle=75, minimum width=1cm, minimum height=1.2cm, right=0.9cm of xa, xshift=(\i-6)*1cm, yshift=0.4cm] (hexagon\i) {};
    }
    
    \foreach \j in {6,7,8} {
        \foreach \i in {0,1,2,3,4,5} {
            \draw[arrow style] (hexagon\i.north) -- (hexagon\j.south);
        }
    }
    
    \end{tikzpicture}}
    }
    \caption{Maya Architecture adapted from LLaVA \cite{liu2024llavanext}}
    \label{fig:onecol}
\end{figure}

\subsection{Pretraining}

For image-text alignment, we used a projection matrix $W$ that brings image features $X_v$ closer to language features. This projection matrix is a simple 2-layer MLP with GELU activation \citep{hendrycks2016gaussian}, as in LLaVA 1.5 \citep{liu2023improvedllava}. Although we experimented with 4- and 8-layer MLPs, the 2-layer configuration consistently achieved the lowest training loss. Advanced alignment techniques, such as gated soft-attention in Flamingo \cite{alayrac2022flamingo}, Q-Former in BLIP-2 \cite{li2023blip}, or pooling from MM1 \cite{mckinzie2024mm1} as alternatives to the projection layer, are set aside for future work. For each image $X_v$, we used the multi-turn conversation data $$(X^1_q,X^1_a, \dots, X^T_q, X^T_a)$$ where $T$ is the total number of turns from LLaVA \cite{liu2023llava}. We translated this data to 7 languages. We pretrained Maya on this dataset and the toxicity free version of the same. Image inputs were cropped to 256x256 for compatibility with the SigLIP encoder. Training was conducted on 8xH100 GPUs with 80GB DRAM, using a per-device batch size of 32 and a global batch size of 256. A learning rate of 1e-3 and a cosine scheduler were applied during training. The pretraining process only trains the projection matrix and took about 20 hours.

\subsection{Finetuning}

We instruction-finetuned our pretrained Maya model using the PALO 150K instruction-tuning dataset \citep{maaz2024palo}. During finetuning, we observed that Low Rank Adaptation (LoRA) \cite{hu2021lora} produced suboptimal results, particularly when both adapter matrices $A$ and $B$ were updated with the same learning rate \cite{hayou2024lora+}. Based on these findings, we opted against using LoRA and instead implemented full finetuning on 8xH100 GPUs. Our finetuning configuration used a per-device batch size of 16 and a global batch size of 128. Finetuning process took about 48 hours.

We kept both the vision encoder and the language encoders frozen during the training process. We did pretraining and finetuning for both versions of our dataset: the pretraining dataset translated into 7 languages, with and without toxicity filtering. This resulted in two variants of our model: Maya and Maya-Toxicity-Free.

\section{Results}
\label{sec:results}
We evaluate Maya on PALO multilingual evaluation set \cite{maaz2024palo} as shown in Table \ref{tab:model-performance}. Although our pretraining dataset only contains eight languages, we finetuned Maya on PALO instruction tuning dataset that contains ten languages. Therefore in Table \ref{tab:model-performance} Maya evaluation results are shown in all of those ten PALO instruction-tuning languages. Maya offers good performance across languages compared to the models within the size class of 7B models, and offers comparable performance to the 13B size models. On average across all languages, Maya outperforms LLaVA-13B and is only slightly worse than PALO-13B. Specifically, among the common eight languages, Maya performs better than PALO 7B in five languages. We think this is due to the fact that Maya pretrain dataset is multilingual in nature whereas PALO pretrain dataset is same as the English-only LLaVA pretrain dataset. Maya also beats PALO and LLaVA in Arabic language for both the size classes of 7B and 13B. The root based language system of Arabic along with the quality of translation due to the choice of preamble led to this result. Another observation we have with PALO multilingual evaluation is that the scores for \textbf{Maya} and \textbf{Maya-Toxicity Free} are exactly the same. We empirically conclude that this is because multilingual evaluation does not account for content toxicity. 

\begin{figure}[h]
    \centering
    \begin{conversationbox}{Close-Up of Multiple Everyday Objects}
        \includegraphics[width=1\linewidth]{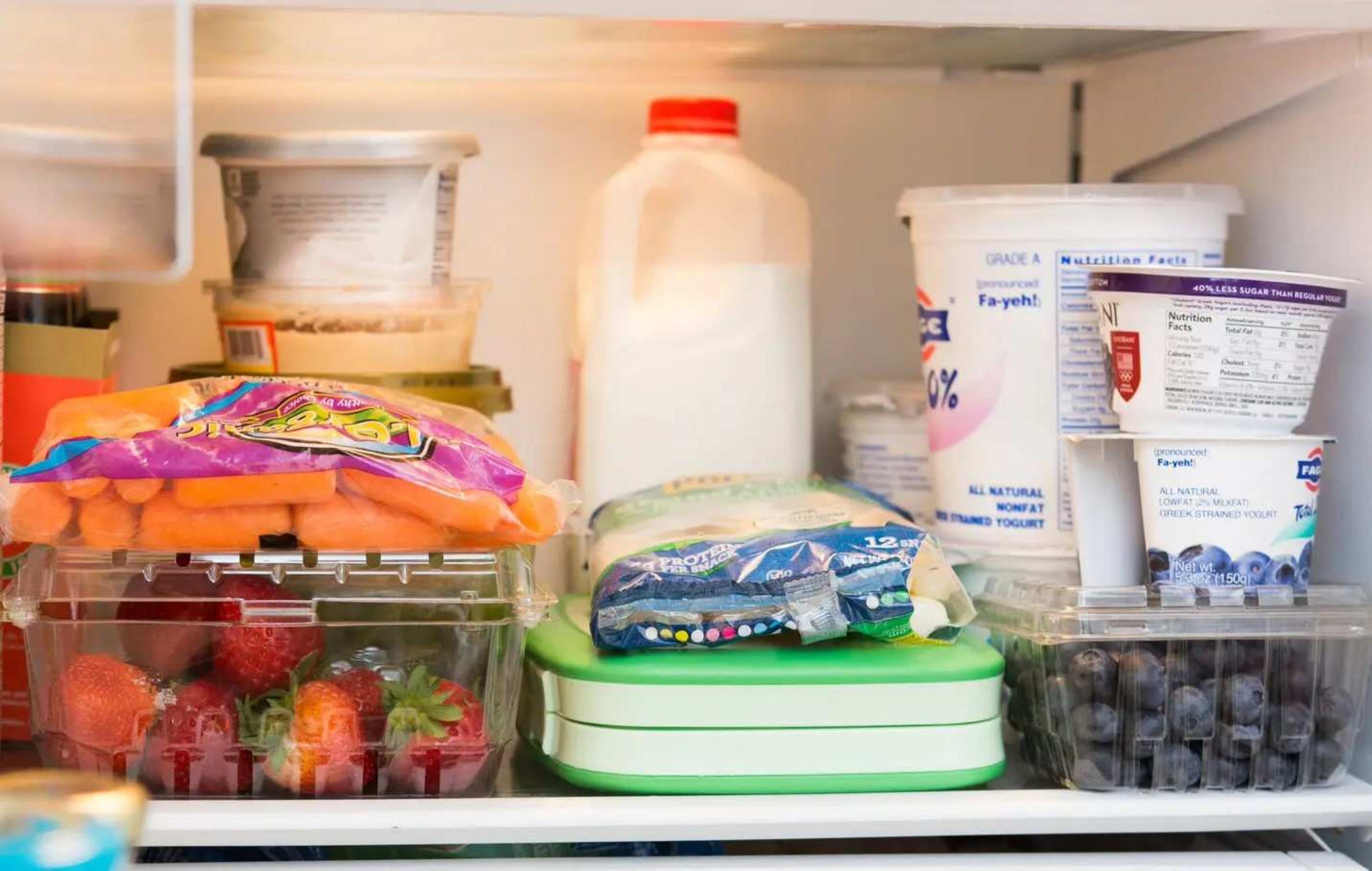}
        
        \textbf{Question 1:} \\
        What fruit is in the left part of the fridge? 
        \tcblower
        \textbf{Maya:} \\
        There are strawberries in the left part of the fridge.\\
        \textbf{LLaVA-7B:} \\
        There are strawberries in the left part of the fridge.\\
        \textbf{GPT4:} \\
        In the left part of the fridge, there is a plastic box of strawberries.
    \end{conversationbox}
    
    \begin{subconversationbox}
        \textbf{Question 2:} \\
        What is the brand of the yogurt flavored with blueberry?
        \tcblower
        \textbf{Maya:} \\
        The brand of the yogurt flavored with blueberry is Yoplait.\\
        \textbf{LLaVA-7B:} \\
        The brand of the yogurt flavored with blueberry is Chobani.\\
        \textbf{GPT4:} \\
        The brand of the blueberry-flavored yogurt is Fage.
    \end{subconversationbox}
    \caption{Comparison of model responses to visual questions on everyday objects}
    \label{fig:model_responses}

\end{figure}

In Figure \ref{fig:model_responses}, we compare Maya responses with LLaVA and GPT4. We observe that Maya responses are very similar to LLaVA and lacks the similar level of details vs. GPT4. For example, in answering \textbf{Question 1}, both Maya and LLaVA lacks the details with respect to \textbf{a plastic box}. When Maya is wrong, for example in \textbf{Question 2}, LLaVA is also wrong; although in a different way. The right answer here is \textbf{Fage} from GPT4. Both Maya and LLaVA get it wrong saying \textbf{Yoplait} and \textbf{Chobani} respectively. Additional qualitative evaluation results are added in the supplemental materials. 

\definecolor{highlight}{gray}{0.9}
\definecolor{neg}{RGB}{0,0,255}
\definecolor{pos}{RGB}{255,0,0}
\begin{table*}[h]
    \centering
    \begin{tabular}{lcccccccccc|c}
        \toprule
        Model & English & Chinese & French & Spanish & Russian & Japanese & Arabic & Hindi & Bengali & Urdu & Avg. \\
        \midrule
        \rowcolor{highlight}\textbf{Maya} (8B) & 61.5 & \underline{61.7} & 61.0 & 60.4 & \underline{62.2} & \underline{63.7} & \textbf{\underline{63.4}} & \underline{64.0} & 50.8 & 55 & \underline{60.4} \\
        LLaVA-7B & \underline{67.9} & 55.7 & \underline{62.4} & \underline{64.5} & 55.3 & 59.2 & 38.9 & 29.4 & 13.9 & 21.8 & 46.9 \\
        PALO-7B & 64.2 & 55.7 & 58.3 & 61.0 & 57.4 & 57.5 & 57.8 & 57.6 & \underline{51.7} & \underline{55.3} & 57.7 \\
        \midrule
        & \textcolor{neg}{-6.4} & \textcolor{pos}{+6.0} & \textcolor{neg}{-1.4} & \textcolor{neg}{-4.1} & \textcolor{pos}{+4.8} & \textcolor{pos}{+6.2} & \textcolor{pos}{+5.6} & 
        \textcolor{pos}{+6.4} & \textcolor{neg}{-0.9} & \textcolor{neg}{-0.3} & \textcolor{pos}{+2.7} \\
        \midrule
        LLaVA-13B & \textbf{\underline{69.5}} & \textbf{\underline{62.9}} & \textbf{\underline{67.5}} & 64.6 & 62.3 & \textbf{\underline{65.3}} & 37.2 & 27.8 & 20.4 & 22.1 & 49.9 \\
        PALO-13B & 65.5 & 62.1 & 66.4 & \textbf{\underline{65.9}} & \textbf{\underline{62.4}} & 60.6 & \underline{56.9} & \textbf{\underline{66.8}} & \textbf{\underline{53.5}} & \textbf{\underline{59.6}} & \textbf{\underline{61.9}} \\
        \midrule
        & \textcolor{neg}{-8.0} & \textcolor{neg}{-1.2} & \textcolor{neg}{-6.5} & \textcolor{neg}{-5.5} & \textcolor{neg}{-0.2} & \textcolor{neg}{-1.6} & \textcolor{pos}{+6.5} & \textcolor{neg}{-2.8} & 
        \textcolor{neg}{-2.7} & \textcolor{neg}{-4.6} & \textcolor{neg}{-1.5} \\
        \bottomrule
    \end{tabular}
    \caption{A comparison of LLaVA and PALO with Maya on eight languages adapted from LLaVA-Bench (In-the-Wild). Values \underline{underlined} indicate best performance within size class and values in \textbf{bold} indicate best performance across all models tested. We provide performance differences between Maya and the best competing model within the size classes where \textcolor{pos}{red} indicates where Maya is performing better and \textcolor{neg}{blue} indicates where Maya is performing worse than the best in the size class. ``Avg." represents the average over all the languages.}
    \label{tab:model-performance}
\end{table*}

\begin{table}[ht!]
\centering
\begin{tabular}{>{\raggedright\arraybackslash}p{0.5\linewidth}p{0.12\linewidth}p{0.12\linewidth}}
\toprule
\textbf{Benchmarks} & \textbf{Maya} & \textbf{Maya-Toxicity Free} \\
\midrule
GQA                        & 57.79\% & 57.84\% \\
VizWiz                     & 34.92\% & 34.98\% \\
ScienceQA                  & 70.27\% & 69.51\% \\
TextVQA                    & 47.01\% & 48.56\% \\
POPE-adversarial           & 81.00\% & 80.97\% \\
POPE-popular               & 84.10\% & 82.84\% \\
POPE-random                & 85.30\% & 84.27\% \\
MMBench                    & 71.10\% & 71.12\% \\
MM-VeT                     & 29.8    & 27.7    \\
MME (Perception + Cognition) & 72.45\% & 71.90\% \\
\multicolumn{3}{l}{MME (Cognition):} \\
\hspace{1em}Commonsense Reasoning      & 75.71\% & 68.57\% \\
\hspace{1em}Numerical Calculation      & 47.50\% & 50.00\% \\
\hspace{1em}Text Translation           & 52.50\% & 55.00\% \\
\hspace{1em}Code Reasoning             & 60.00\% & 57.50\% \\
\bottomrule
\end{tabular}
\caption{Accuracy of Maya models on English Language across multiple benchmarks.}
\label{table:maya_eval_english}
\end{table}


\begin{table}[ht!]
\centering
\begin{tabular}{lcc}
\toprule
\textbf{VizWiz} & \textbf{Maya} & \textbf{Maya-Toxicity Free} \\
\midrule
\textbf{Overall}              & \textbf{34.92\% }& \textbf{34.98\% }\\
\midrule
other         & 34.03\% & 33.66\% \\
unanswerable  & 30.88\% & 32.03\% \\
yes/no        & 77.02\% & 77.08\% \\
number        & 24.63\% & 24.72\% \\
\bottomrule
\end{tabular}
\caption{Detailed accuracy results for VizWiz.}
\label{table:vizwiz_results}
\end{table}

\begin{table}[ht!]
\centering
\begin{tabular}{>{\raggedright\arraybackslash}p{0.6\linewidth}p{0.12\linewidth}p{0.12\linewidth}}
\toprule
\textbf{MMVeT} & \textbf{Maya} & \textbf{Maya-Toxicity Free} \\
\midrule
\textbf{Overall}              & \textbf{29.8} & \textbf{27.7} \\
\midrule
\multicolumn{3}{c}{\textbf{Capabilities}} \\
\midrule
Recognition          & 33.7 & 32.7 \\
OCR                  & 23.1 & 18.7 \\
Knowledge            & 16.0 & 13.7 \\
Language Generation  & 18.4 & 19.3 \\
Spatial Awareness    & 29.0 & 26.7 \\
Math                 & 15.4 & 7.7  \\
\midrule
\multicolumn{3}{c}{\textbf{Capability Integrations}} \\
\midrule
Language Generation, Recognition, Knowledge   & 15.5 & 16.5 \\
Recognition                                  & 69.2 & 68.9 \\
Spatial Awareness, OCR                       & 25.8 & 25.4 \\
Spatial Awareness, OCR, Math                 & 28.6 & 14.3 \\
Spatial Awareness, Recognition               & 41.7 & 50.0 \\
OCR                                          & 25.8 & 18.3 \\
OCR, Math                                    & 0.0  & 0.0  \\
Recognition, Knowledge                       & 16.7 & 6.0  \\
Spatial Awareness, Language Generation, OCR, Recognition & 44.2 & 49.8 \\
Language Generation, OCR, Recognition, Knowledge        & 15.2 & 9.0  \\
Spatial Awareness, OCR, Recognition          & 14.3 & 14.3 \\
OCR, Recognition                             & 53.0 & 25.0 \\
Spatial Awareness, OCR, Knowledge            & 3.3  & 0.0  \\
Spatial Awareness, Recognition, Knowledge    & 50.0 & 0.0  \\
Spatial Awareness, Language Generation, OCR  & 19.0 & 24.0 \\
Spatial Awareness, OCR, Recognition, Math    & 0.0  & 0.0  \\
\bottomrule
\end{tabular}
\caption{MMVeT results including individual and integrated capabilities. Values are mean of 5 runs with 0.2 standard deviation.}
\label{table:mmvet_results}
\end{table}

To compare the effect of filtering the pretraining dataset for toxicity, we evaluate Maya and Maya-Toxicity Free models across various English-only benchmarks as shown in Table \ref{table:maya_eval_english}. Both models have comparable accuracy across most of the benchmarks, with only slight variations. This suggests that removing toxic content from the training data does not significantly degrade overall performance for these tasks. Maya-Toxicity Free shows marginal gains in TextVQA, Text Translation, and Numerical Calculation benchmarks. Performance in Commonsense Reasoning and MM-VeT decreases, indicating that some complex reasoning tasks may benefit from the presence of diverse (and possibly toxic) content in training data.

In Table \ref{table:vizwiz_results}, we show how Maya compares with its toxicity free variant in various VizWiz categories. Maya-Toxicity Free slightly outperforms Maya by 0.06\%. This indicates that the removal of toxic content in training data has a minimal positive impact on the overall performance of the VizWiz benchmark. Maya performs slightly better in the “other” category. This suggests that for questions that fall outside of well-defined categories, the unfiltered training data may offer a slight advantage. Maya-Toxicity Free shows improved performance (32.03\% vs. 30.88\%), suggesting that a cleaner training dataset helps the model better recognize when a question cannot be answered. Both models perform almost identically in the yes/no category indicating that toxicity removal does not significantly affect binary question accuracy.

In Table \ref{table:mmvet_results}, we show a detailed performance analysis of the Maya and Maya-Toxicity Free models on the MMVeT benchmark, breaking down their overall accuracy, individual capabilities, and more complex capability integrations. In categories like \textbf{Spatial Awareness, OCR, Knowledge} and \textbf{Spatial Awareness, Language Generation, OCR}, Maya-Toxicity Free shows zero performance, suggesting areas where toxicity removal could have eliminated beneficial patterns. For \textbf{Spatial Awareness, OCR, Recognition, Math}, both models perform poorly, indicating the difficulty of this capability combination, with no difference between the models. Overall, MMVeT results highlight that while Maya-Toxicity Free shows some improvements in specific areas, like language generation and certain integrations involving spatial awareness, it generally suffers a performance decline, especially in math, OCR, and integrated recognition tasks. The impact of removing toxic content seems to weaken the model's overall ability to perform complex reasoning and integrate multiple capabilities effectively. We leave out further investigation on this topic for future work.

\begin{figure}[t]
  \centering
  \fbox{\includegraphics[width=1\linewidth]{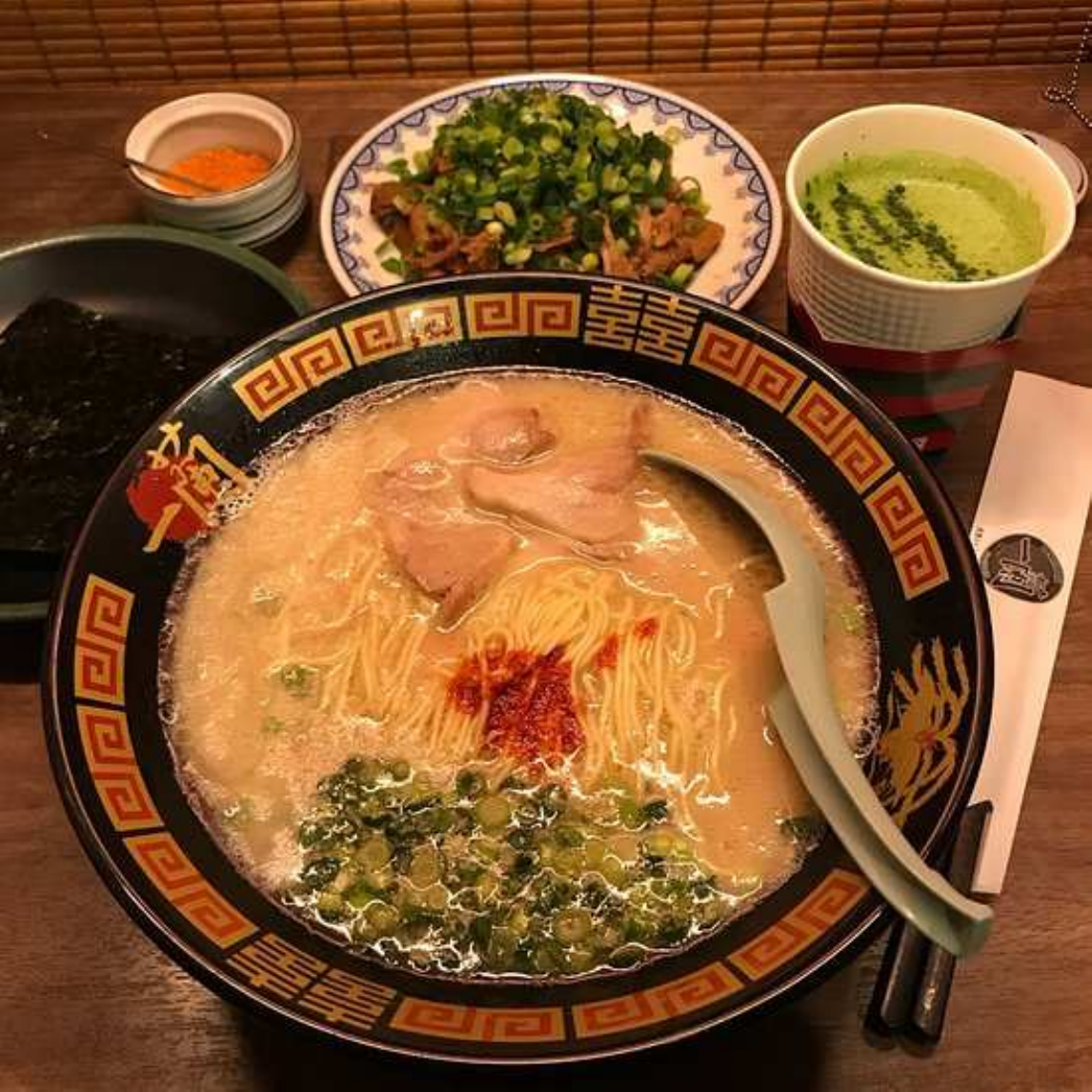}}
  \caption{Example image from LLaVA-Bench (In-the-Wild) \citep{liu2023llava}.}
  \label{fig:asianfood}
\end{figure}

\begin{figure}[t]
  \centering
  \fbox{\includegraphics[width=1\linewidth]{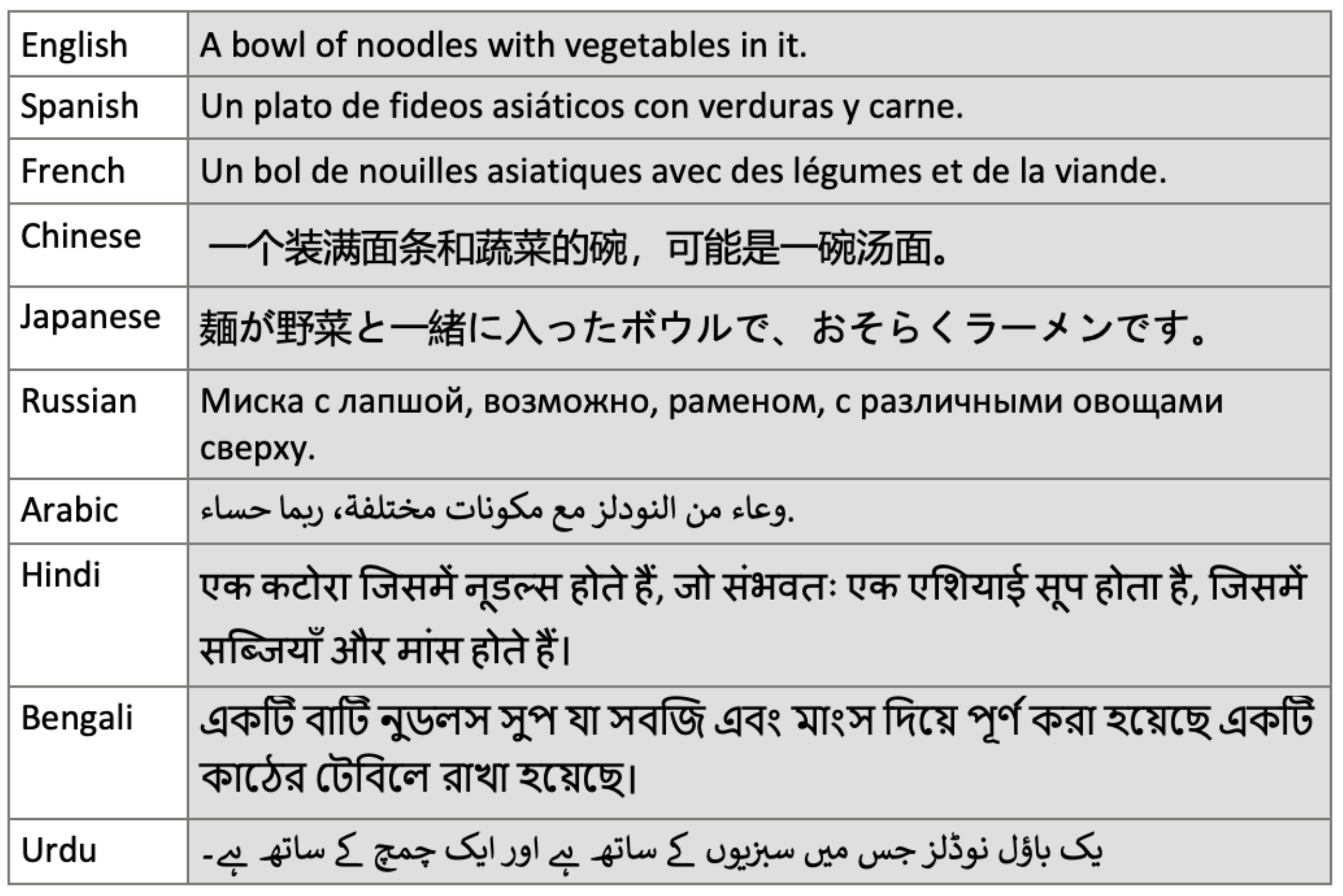}}
  \caption{Maya output for prompt (with image from Figure \ref{fig:asianfood}): Please describe the food in \{language\} in 1 sentence.}
  \label{fig:multilingual_output}
\end{figure}

Figure \ref{fig:multilingual_output} shows visual question output in different languages for figure \ref{fig:asianfood}. We asked Maya \textbf{please describe the food in \{language\} in 1 sentence}. We observe that Bengali reply in this case is more detailed than English, identifying meat on the bowl and wooden table. Spanish, French and Hindi outputs are more detail than English as all three languages identify meat but fails to identify the wooden table as in Bengali. Chinese and Japanese outputs are similar to English. 
\section{Conclusion}
\label{sec:conclusion}

Maya drives high-quality AI content generation across languages and regions, leveraging multilingual, multimodal data to fill gaps, especially for low-resource languages. To ensure safe deployment, we rigorously curate data to remove harmful content, though some residual traces may persist. Future work will focus on refining Maya’s adaptability, including testing alternative projection layers for improved cross-modal alignment and unfreezing decoder layers to optimize layer-specific fine-tuning. We plan to expand our pretrain dataset to include Bengali and Urdu translations and grow the instruction-tuning dataset to 665K examples, enhancing instruction accuracy across languages and modalities. To improve translation quality, we will tailor preambles per language. Additionally, we aim to benchmark rigorously on benchmarks such as PangeaBench \cite{yue2024pangea}, CVQA \citep{romero2024cvqa} etc. to ensure robust, diverse user support.

{
    \small
    \bibliographystyle{ieeenat_fullname}
    \bibliography{main}

\begin{thebibliography}{64}
\providecommand{\natexlab}[1]{#1}
\providecommand{\url}[1]{\texttt{#1}}
\expandafter\ifx\csname urlstyle\endcsname\relax
  \providecommand{\doi}[1]{doi: #1}\else
  \providecommand{\doi}{doi: \begingroup \urlstyle{rm}\Url}\fi

\bibitem[Abdin et~al.(2024)Abdin, Jacobs, Awan, Aneja, Awadallah, Awadalla, Bach, Bahree, Bakhtiari, Behl, et~al.]{abdin2024phi}
Marah Abdin, Sam~Ade Jacobs, Ammar~Ahmad Awan, Jyoti Aneja, Ahmed Awadallah, Hany Awadalla, Nguyen Bach, Amit Bahree, Arash Bakhtiari, Harkirat Behl, et~al.
\newblock {Phi-3 Technical Report: A Highly Capable Language Model Locally on Your Phone}.
\newblock \emph{arXiv preprint arXiv:2404.14219}, 2024.

\bibitem[Achiam et~al.(2023)Achiam, Adler, Agarwal, Ahmad, Akkaya, Aleman, Almeida, Altenschmidt, Altman, Anadkat, et~al.]{achiam2023gpt}
Josh Achiam, Steven Adler, Sandhini Agarwal, Lama Ahmad, Ilge Akkaya, Florencia~Leoni Aleman, Diogo Almeida, Janko Altenschmidt, Sam Altman, Shyamal Anadkat, et~al.
\newblock {GPT-4 Technical Report}.
\newblock \emph{arXiv preprint arXiv:2303.08774}, 2023.

\bibitem[Agrawal et~al.(2024)Agrawal, Antoniak, Hanna, Chaplot, Chudnovsky, Garg, Gervet, Ghosh, H{\'e}liou, Jacob, et~al.]{agrawal2024pixtral}
Pravesh Agrawal, Szymon Antoniak, Emma~Bou Hanna, Devendra Chaplot, Jessica Chudnovsky, Saurabh Garg, Theophile Gervet, Soham Ghosh, Am{\'e}lie H{\'e}liou, Paul Jacob, et~al.
\newblock Pixtral 12b.
\newblock \emph{arXiv preprint arXiv:2410.07073}, 2024.

\bibitem[AI(2024)]{ministral}
Mistral AI.
\newblock {Un Ministral, des Ministraux}, 2024.

\bibitem[Alayrac et~al.(2022)Alayrac, Donahue, Luc, Miech, Barr, Hasson, Lenc, Mensch, Millican, Reynolds, et~al.]{alayrac2022flamingo}
Jean-Baptiste Alayrac, Jeff Donahue, Pauline Luc, Antoine Miech, Iain Barr, Yana Hasson, Karel Lenc, Arthur Mensch, Katherine Millican, Malcolm Reynolds, et~al.
\newblock {Flamingo: a Visual Language Model for Few-Shot Learning}.
\newblock \emph{Advances in Neural Information Processing Systems}, 35:\penalty0 23716--23736, 2022.

\bibitem[Anthropic(2024)]{claude}
Anthropic.
\newblock {Introducing the next generation of Claude}, 2024.

\bibitem[Aryabumi et~al.(2024)Aryabumi, Dang, Talupuru, Dash, Cairuz, Lin, Venkitesh, Smith, Marchisio, Ruder, et~al.]{aryabumi2024aya}
Viraat Aryabumi, John Dang, Dwarak Talupuru, Saurabh Dash, David Cairuz, Hangyu Lin, Bharat Venkitesh, Madeline Smith, Kelly Marchisio, Sebastian Ruder, et~al.
\newblock {Aya 23: Open Weight Releases to Further Multilingual Progress}.
\newblock \emph{arXiv preprint arXiv:2405.15032}, 2024.

\bibitem[Awadalla et~al.(2024)Awadalla, Xue, Shu, Yan, Wang, Purushwalkam, Shen, Lee, Lo, Park, Guha, Savarese, Schmidt, Choi, Xiong, and Xu]{awadalla2024blip3kaleknowledgeaugmentedlargescale}
Anas Awadalla, Le Xue, Manli Shu, An Yan, Jun Wang, Senthil Purushwalkam, Sheng Shen, Hannah Lee, Oscar Lo, Jae~Sung Park, Etash Guha, Silvio Savarese, Ludwig Schmidt, Yejin Choi, Caiming Xiong, and Ran Xu.
\newblock Blip3-kale: Knowledge augmented large-scale dense captions, 2024.

\bibitem[Bai et~al.(2023)Bai, Bai, Chu, Cui, Dang, Deng, Fan, Ge, Han, Huang, Hui, Ji, Li, Lin, Lin, Liu, Liu, Lu, Lu, Ma, Men, Ren, Ren, Tan, Tan, Tu, Wang, Wang, Wang, Wu, Xu, Xu, Yang, Yang, Yang, Yang, Yao, Yu, Yuan, Yuan, Zhang, Zhang, Zhang, Zhang, Zhou, Zhou, Zhou, and Zhu]{qwen}
Jinze Bai, Shuai Bai, Yunfei Chu, Zeyu Cui, Kai Dang, Xiaodong Deng, Yang Fan, Wenbin Ge, Yu Han, Fei Huang, Binyuan Hui, Luo Ji, Mei Li, Junyang Lin, Runji Lin, Dayiheng Liu, Gao Liu, Chengqiang Lu, Keming Lu, Jianxin Ma, Rui Men, Xingzhang Ren, Xuancheng Ren, Chuanqi Tan, Sinan Tan, Jianhong Tu, Peng Wang, Shijie Wang, Wei Wang, Shengguang Wu, Benfeng Xu, Jin Xu, An Yang, Hao Yang, Jian Yang, Shusheng Yang, Yang Yao, Bowen Yu, Hongyi Yuan, Zheng Yuan, Jianwei Zhang, Xingxuan Zhang, Yichang Zhang, Zhenru Zhang, Chang Zhou, Jingren Zhou, Xiaohuan Zhou, and Tianhang Zhu.
\newblock {Qwen Technical Report}.
\newblock \emph{arXiv preprint arXiv:2309.16609}, 2023.

\bibitem[Beyer et~al.(2024)Beyer, Steiner, Pinto, Kolesnikov, Wang, Salz, Neumann, Alabdulmohsin, Tschannen, Bugliarello, et~al.]{beyer2024paligemma}
Lucas Beyer, Andreas Steiner, Andr{\'e}~Susano Pinto, Alexander Kolesnikov, Xiao Wang, Daniel Salz, Maxim Neumann, Ibrahim Alabdulmohsin, Michael Tschannen, Emanuele Bugliarello, et~al.
\newblock {PaliGemma: A versatile 3B VLM for transfer}.
\newblock \emph{arXiv preprint arXiv:2407.07726}, 2024.

\bibitem[Brill et~al.(1998)Brill, Florian, Henderson, and Mangu]{brill-etal-1998-beyond}
Eric Brill, Radu Florian, John~C. Henderson, and Lidia Mangu.
\newblock Beyond n-grams: Can linguistic sophistication improve language modeling?
\newblock In \emph{{COLING} 1998 Volume 1: The 17th International Conference on Computational Linguistics}, 1998.

\bibitem[Bugliarello et~al.(2022)Bugliarello, Liu, Pfeiffer, Reddy, Elliott, Ponti, and Vuli{\'c}]{bugliarello-etal-2022-iglue}
Emanuele Bugliarello, Fangyu Liu, Jonas Pfeiffer, Siva Reddy, Desmond Elliott, Edoardo~Maria Ponti, and Ivan Vuli{\'c}.
\newblock {IGLUE}: A benchmark for transfer learning across modalities, tasks, and languages.
\newblock In \emph{Proceedings of the 39th International Conference on Machine Learning}, pages 2370--2392. PMLR, 2022.

\bibitem[Chen et~al.(2022)Chen, Wang, Changpinyo, Piergiovanni, Padlewski, Salz, Goodman, Grycner, Mustafa, Beyer, et~al.]{chen2022pali}
Xi Chen, Xiao Wang, Soravit Changpinyo, AJ Piergiovanni, Piotr Padlewski, Daniel Salz, Sebastian Goodman, Adam Grycner, Basil Mustafa, Lucas Beyer, et~al.
\newblock {PaLI: A Jointly-Scaled Multilingual Language-Image Model}.
\newblock \emph{arXiv preprint arXiv:2209.06794}, 2022.

\bibitem[Chen et~al.(2023)Chen, Djolonga, Padlewski, Mustafa, Changpinyo, Wu, Ruiz, Goodman, Wang, Tay, et~al.]{chen2023pali}
Xi Chen, Josip Djolonga, Piotr Padlewski, Basil Mustafa, Soravit Changpinyo, Jialin Wu, Carlos~Riquelme Ruiz, Sebastian Goodman, Xiao Wang, Yi Tay, et~al.
\newblock {PaLI-X: On Scaling up a Multilingual Vision and Language Model}.
\newblock \emph{arXiv preprint arXiv:2305.18565}, 2023.

\bibitem[Cohere(2024)]{command_r}
Cohere.
\newblock {Command R}.
\newblock https://cohere.com/command, 2024.

\bibitem[Deitke et~al.(2024)Deitke, Clark, Lee, Tripathi, Yang, Park, Salehi, Muennighoff, Lo, Soldaini, et~al.]{deitke2024molmo}
Matt Deitke, Christopher Clark, Sangho Lee, Rohun Tripathi, Yue Yang, Jae~Sung Park, Mohammadreza Salehi, Niklas Muennighoff, Kyle Lo, Luca Soldaini, et~al.
\newblock {Molmo and PixMo: Open Weights and Open Data for State-of-the-Art Multimodal Models}.
\newblock \emph{arXiv preprint arXiv:2409.17146}, 2024.

\bibitem[Dubey et~al.(2024)Dubey, Jauhri, Pandey, Kadian, Al-Dahle, Letman, Mathur, Schelten, Yang, Fan, et~al.]{dubey2024llama}
Abhimanyu Dubey, Abhinav Jauhri, Abhinav Pandey, Abhishek Kadian, Ahmad Al-Dahle, Aiesha Letman, Akhil Mathur, Alan Schelten, Amy Yang, Angela Fan, et~al.
\newblock {The Llama 3 Herd of Models}.
\newblock \emph{arXiv preprint arXiv:2407.21783}, 2024.

\bibitem[Elliott et~al.(2016)Elliott, Frank, Sima{'}an, and Specia]{elliott2016multi30k}
Desmond Elliott, Stella Frank, Khalil Sima{'}an, and Lucia Specia.
\newblock {M}ulti30{K}: Multilingual {E}nglish-{G}erman image descriptions.
\newblock In \emph{Proceedings of the 5th Workshop on Vision and Language}, pages 70--74, Berlin, Germany, 2016. Association for Computational Linguistics.

\bibitem[Geigle et~al.(2023)Geigle, Jain, Timofte, and Glava{\v{s}}]{geigle2023mblip}
Gregor Geigle, Abhay Jain, Radu Timofte, and Goran Glava{\v{s}}.
\newblock {mBLIP: Efficient Bootstrapping of Multilingual Vision-LLMs}.
\newblock \emph{arXiv preprint arXiv:2307.06930}, 2023.

\bibitem[Hanu and {Unitary team}(2020)]{Detoxify}
Laura Hanu and {Unitary team}.
\newblock Detoxify.
\newblock Github. https://github.com/unitaryai/detoxify, 2020.

\bibitem[Hayou et~al.(2024)Hayou, Ghosh, and Yu]{hayou2024lora+}
Soufiane Hayou, Nikhil Ghosh, and Bin Yu.
\newblock {LoRA+: Efficient Low Rank Adaptation of Large Models}.
\newblock \emph{arXiv preprint arXiv:2402.12354}, 2024.

\bibitem[Helff et~al.(2024)Helff, Friedrich, Brack, Kersting, and Schramowski]{helff2024llavaguard}
Lukas Helff, Felix Friedrich, Manuel Brack, Kristian Kersting, and Patrick Schramowski.
\newblock {LLavaGuard: VLM-based Safeguards for Vision Dataset Curation and Safety Assessment}.
\newblock \emph{arXiv preprint arXiv:2406.05113}, 2024.

\bibitem[Hendrycks and Gimpel(2016)]{hendrycks2016gaussian}
Dan Hendrycks and Kevin Gimpel.
\newblock {Gaussian Error Linear Units (GELUs)}.
\newblock \emph{arXiv preprint arXiv:1606.08415}, 2016.

\bibitem[Hinck et~al.(2024)Hinck, Holtermann, Olson, Schneider, Yu, Bhiwandiwalla, Lauscher, Tseng, and Lal]{hinck2024llava}
Musashi Hinck, Carolin Holtermann, Matthew~Lyle Olson, Florian Schneider, Sungduk Yu, Anahita Bhiwandiwalla, Anne Lauscher, Shaoyen Tseng, and Vasudev Lal.
\newblock {Why do LLaVA Vision-Language Models Reply to Images in English?}
\newblock \emph{arXiv preprint arXiv:2407.02333}, 2024.

\bibitem[Hu et~al.(2021)Hu, Shen, Wallis, Allen-Zhu, Li, Wang, Wang, and Chen]{hu2021lora}
Edward~J Hu, Yelong Shen, Phillip Wallis, Zeyuan Allen-Zhu, Yuanzhi Li, Shean Wang, Lu Wang, and Weizhu Chen.
\newblock {LoRA: Low-Rank Adaptation of Large Language Models}.
\newblock \emph{arXiv preprint arXiv:2106.09685}, 2021.

\bibitem[Joshi et~al.(2020)Joshi, Santy, Budhiraja, Bali, and Choudhury]{joshi2020state}
Pratik Joshi, Sebastin Santy, Amar Budhiraja, Kalika Bali, and Monojit Choudhury.
\newblock {The State and Fate of Linguistic Diversity and Inclusion in the NLP World}.
\newblock In \emph{Proceedings of the 58th Annual Meeting of the Association for Computational Linguistics}, pages 6282--6293, Online, 2020. Association for Computational Linguistics.

\bibitem[Krishna et~al.(2017)Krishna, Zhu, Groth, Johnson, Hata, Kravitz, Chen, Kalantidis, Li, Shamma, et~al.]{krishna2017visual}
Ranjay Krishna, Yuke Zhu, Oliver Groth, Justin Johnson, Kenji Hata, Joshua Kravitz, Stephanie Chen, Yannis Kalantidis, Li-Jia Li, David~A Shamma, et~al.
\newblock {Visual Genome: Connecting Language and Vision Using Crowdsourced Dense Image Annotations}.
\newblock \emph{International journal of computer vision}, 123:\penalty0 32--73, 2017.

\bibitem[Le~Scao et~al.(2023)Le~Scao, Fan, Akiki, Pavlick, Ili{\'c}, Hesslow, Castagn{\'e}, Luccioni, Yvon, Gall{\'e}, et~al.]{le2023bloom}
Teven Le~Scao, Angela Fan, Christopher Akiki, Ellie Pavlick, Suzana Ili{\'c}, Daniel Hesslow, Roman Castagn{\'e}, Alexandra~Sasha Luccioni, Fran{\c{c}}ois Yvon, Matthias Gall{\'e}, et~al.
\newblock {BLOOM: A 176B-Parameter Open-Access Multilingual Language Model}.
\newblock \emph{arXiv preprint arXiv:2211.05100}, 2023.

\bibitem[Li et~al.(2023)Li, Li, Savarese, and Hoi]{li2023blip}
Junnan Li, Dongxu Li, Silvio Savarese, and Steven Hoi.
\newblock {BLIP-2: Bootstrapping Language-Image Pre-training with Frozen Image Encoders and Large Language Models}.
\newblock In \emph{International conference on machine learning}, pages 19730--19742. PMLR, 2023.

\bibitem[Lin et~al.(2024)Lin, Ji, Tiedemann, Martins, and Sch{\"u}tze]{lin2024mala}
Peiqin Lin, Shaoxiong Ji, J{\"o}rg Tiedemann, Andr{\'e}~FT Martins, and Hinrich Sch{\"u}tze.
\newblock {MaLA-500: Massive Language Adaptation of Large Language Models}.
\newblock \emph{arXiv preprint arXiv:2401.13303}, 2024.

\bibitem[Lin et~al.(2014)Lin, Maire, Belongie, Hays, Perona, Ramanan, Doll{\'a}r, and Zitnick]{lin2014microsoft}
Tsung-Yi Lin, Michael Maire, Serge Belongie, James Hays, Pietro Perona, Deva Ramanan, Piotr Doll{\'a}r, and C~Lawrence Zitnick.
\newblock {Microsoft COCO: Common Objects in Context}.
\newblock In \emph{Computer Vision--ECCV 2014: 13th European Conference, Zurich, Switzerland, September 6-12, 2014, Proceedings, Part V 13}, pages 740--755. Springer, 2014.

\bibitem[Lin et~al.(2021)Lin, Mihaylov, Artetxe, Wang, Chen, Simig, Ott, Goyal, Bhosale, Du, et~al.]{lin2021few}
Xi~Victoria Lin, Todor Mihaylov, Mikel Artetxe, Tianlu Wang, Shuohui Chen, Daniel Simig, Myle Ott, Naman Goyal, Shruti Bhosale, Jingfei Du, et~al.
\newblock Few-shot learning with multilingual language models.
\newblock \emph{arXiv preprint arXiv:2112.10668}, 2021.

\bibitem[Liu et~al.(2021)Liu, Bugliarello, Ponti, Reddy, Collier, and Elliott]{liu-etal-2021-visually}
Fangyu Liu, Emanuele Bugliarello, Edoardo~Maria Ponti, Siva Reddy, Nigel Collier, and Desmond Elliott.
\newblock {Visually Grounded Reasoning across Languages and Cultures}.
\newblock In \emph{Proceedings of the 2021 Conference on Empirical Methods in Natural Language Processing}, pages 10467--10485, Online and Punta Cana, Dominican Republic, 2021. Association for Computational Linguistics.

\bibitem[Liu et~al.(2023{\natexlab{a}})Liu, Li, Li, and Lee]{liu2023improvedllava}
Haotian Liu, Chunyuan Li, Yuheng Li, and Yong~Jae Lee.
\newblock {Improved Baselines with Visual Instruction Tuning}, 2023{\natexlab{a}}.

\bibitem[Liu et~al.(2023{\natexlab{b}})Liu, Li, Wu, and Lee]{liu2023llava}
Haotian Liu, Chunyuan Li, Qingyang Wu, and Yong~Jae Lee.
\newblock {Visual Instruction Tuning}, 2023{\natexlab{b}}.

\bibitem[Liu et~al.(2024)Liu, Li, Li, Li, Zhang, Shen, and Lee]{liu2024llavanext}
Haotian Liu, Chunyuan Li, Yuheng Li, Bo Li, Yuanhan Zhang, Sheng Shen, and Yong~Jae Lee.
\newblock {LLaVA-NeXT: Improved reasoning, OCR, and world knowledge}, 2024.

\bibitem[Maaz et~al.(2024)Maaz, Rasheed, Shaker, Khan, Cholakal, Anwer, Baldwin, Felsberg, and Khan]{maaz2024palo}
Muhammad Maaz, Hanoona Rasheed, Abdelrahman Shaker, Salman Khan, Hisham Cholakal, Rao~M Anwer, Tim Baldwin, Michael Felsberg, and Fahad~S Khan.
\newblock {PALO: A Polyglot Large Multimodal Model for 5B People}.
\newblock \emph{arXiv preprint arXiv:2402.14818}, 2024.

\bibitem[McKinzie et~al.(2024)McKinzie, Gan, Fauconnier, Dodge, Zhang, Dufter, Shah, Du, Peng, Weers, et~al.]{mckinzie2024mm1}
Brandon McKinzie, Zhe Gan, Jean-Philippe Fauconnier, Sam Dodge, Bowen Zhang, Philipp Dufter, Dhruti Shah, Xianzhi Du, Futang Peng, Floris Weers, et~al.
\newblock {MM1: Methods, Analysis \& Insights from Multimodal LLM Pre-training}.
\newblock \emph{arXiv preprint arXiv:2403.09611}, 2024.

\bibitem[OpenAI(2024)]{gpt4o}
OpenAI.
\newblock {Hello GPT-4o}, 2024.

\bibitem[Pan et~al.(2023)Pan, Dong, Huang, Peng, Chen, and Wei]{kosmos-g}
Xichen Pan, Li Dong, Shaohan Huang, Zhiliang Peng, Wenhu Chen, and Furu Wei.
\newblock {Kosmos-G: Generating Images in Context with Multimodal Large Language Models}.
\newblock \emph{ArXiv}, abs/2310.02992, 2023.

\bibitem[Papineni et~al.(2002)Papineni, Roukos, Ward, and Zhu]{papineni2002bleu}
Kishore Papineni, Salim Roukos, Todd Ward, and Wei-Jing Zhu.
\newblock {BLEU: a Method for Automatic Evaluation of Machine Translation}.
\newblock In \emph{Proceedings of the 40th annual meeting of the Association for Computational Linguistics}, pages 311--318, 2002.

\bibitem[Peng et~al.(2023)Peng, Wang, Dong, Hao, Huang, Ma, and Wei]{peng2023kosmos}
Zhiliang Peng, Wenhui Wang, Li Dong, Yaru Hao, Shaohan Huang, Shuming Ma, and Furu Wei.
\newblock {Kosmos-2: Grounding Multimodal Large Language Models to the World}.
\newblock \emph{arXiv preprint arXiv:2306.14824}, 2023.

\bibitem[Pfeiffer et~al.(2022)Pfeiffer, Geigle, Kamath, Steitz, Roth, Vuli{\'c}, and Gurevych]{pfeiffer2021xgqa}
Jonas Pfeiffer, Gregor Geigle, Aishwarya Kamath, Jan-Martin~O. Steitz, Stefan Roth, Ivan Vuli{\'c}, and Iryna Gurevych.
\newblock {x{GQA}: Cross-Lingual Visual Question Answering}.
\newblock In \emph{Findings of the Association for Computational Linguistics: ACL 2022}, pages 2497--2511, Dublin, Ireland, 2022. Association for Computational Linguistics.

\bibitem[Radford et~al.(2021)Radford, Kim, Hallacy, Ramesh, Goh, Agarwal, Sastry, Askell, Mishkin, Clark, et~al.]{radford2021learning}
Alec Radford, Jong~Wook Kim, Chris Hallacy, Aditya Ramesh, Gabriel Goh, Sandhini Agarwal, Girish Sastry, Amanda Askell, Pamela Mishkin, Jack Clark, et~al.
\newblock {Learning Transferable Visual Models From Natural Language Supervision}.
\newblock In \emph{International conference on machine learning}, pages 8748--8763. PMLR, 2021.

\bibitem[Romero et~al.(2024)Romero, Lyu, Wibowo, Lynn, Hamed, Kishore, Mandal, Dragonetti, Abzaliev, Tonja, et~al.]{romero2024cvqa}
David Romero, Chenyang Lyu, Haryo~Akbarianto Wibowo, Teresa Lynn, Injy Hamed, Aditya~Nanda Kishore, Aishik Mandal, Alina Dragonetti, Artem Abzaliev, Atnafu~Lambebo Tonja, et~al.
\newblock {CVQA: Culturally-diverse Multilingual Visual Question Answering Benchmark}.
\newblock \emph{arXiv preprint arXiv:2406.05967}, 2024.

\bibitem[Schuhmann et~al.(2022)Schuhmann, Beaumont, Vencu, Gordon, Wightman, Cherti, Coombes, Katta, Mullis, Wortsman, et~al.]{schuhmann2022laion}
Christoph Schuhmann, Romain Beaumont, Richard Vencu, Cade Gordon, Ross Wightman, Mehdi Cherti, Theo Coombes, Aarush Katta, Clayton Mullis, Mitchell Wortsman, et~al.
\newblock {LAION-5B: An open large-scale dataset for training next generation image-text models}.
\newblock \emph{Advances in Neural Information Processing Systems}, 35:\penalty0 25278--25294, 2022.

\bibitem[Shannon(1948)]{shannon1948mathematical}
Claude~Elwood Shannon.
\newblock A mathematical theory of communication.
\newblock \emph{The Bell system technical journal}, 27\penalty0 (3):\penalty0 379--423, 1948.

\bibitem[Shin et~al.(2024)Shin, Lim, Won, Choi, Kim, Song, Yoo, Kim, and Lim]{shin2024x}
Dongjae Shin, HyeonSeok Lim, Inho Won, Changsu Choi, Minjun Kim, Seungwoo Song, Hangyeol Yoo, Sangmin Kim, and Kyungtae Lim.
\newblock {X-LLaVA: Optimizing Bilingual Large Vision-Language Alignment}.
\newblock \emph{arXiv preprint arXiv:2403.11399}, 2024.

\bibitem[Srinivasan et~al.(2021)Srinivasan, Raman, Chen, Bendersky, and Najork]{wit_2021}
Krishna Srinivasan, Karthik Raman, Jiecao Chen, Michael Bendersky, and Marc Najork.
\newblock {WIT: Wikipedia-Based Image Text Dataset for Multimodal Multilingual Machine Learning}.
\newblock In \emph{Proceedings of the 44th International ACM SIGIR Conference on Research and Development in Information Retrieval}, page 2443–2449, New York, NY, USA, 2021. Association for Computing Machinery.

\bibitem[Su et~al.(2021)Su, Lu, Pan, Wen, and Liu]{su2021roformer}
Jianlin Su, Yu Lu, Shengfeng Pan, Bo Wen, and Yunfeng Liu.
\newblock {RoFormer: Enhanced Transformer with Rotary Position Embedding}, 2021.

\bibitem[Sun et~al.(2024)Sun, Zhou, Li, Lu, Yi, Chen, Xu, Luo, Zhang, Zhan, and Ye]{sun2024parrotmultilingualvisualinstruction}
Hai-Long Sun, Da-Wei Zhou, Yang Li, Shiyin Lu, Chao Yi, Qing-Guo Chen, Zhao Xu, Weihua Luo, Kaifu Zhang, De-Chuan Zhan, and Han-Jia Ye.
\newblock {Parrot: Multilingual Visual Instruction Tuning}, 2024.

\bibitem[Tang et~al.(2024)Tang, Liu, Ye, Lu, Wei, Lin, Li, Mahmood, Feng, Zhao, et~al.]{tang2024mtvqa}
Jingqun Tang, Qi Liu, Yongjie Ye, Jinghui Lu, Shu Wei, Chunhui Lin, Wanqing Li, Mohamad Fitri Faiz~Bin Mahmood, Hao Feng, Zhen Zhao, et~al.
\newblock {MTVQA: Benchmarking Multilingual Text-Centric Visual Question Answering}.
\newblock \emph{arXiv preprint arXiv:2405.11985}, 2024.

\bibitem[Team et~al.(2023)Team, Anil, Borgeaud, Alayrac, Yu, Soricut, Schalkwyk, Dai, Hauth, Millican, et~al.]{team2023gemini}
Gemini Team, Rohan Anil, Sebastian Borgeaud, Jean-Baptiste Alayrac, Jiahui Yu, Radu Soricut, Johan Schalkwyk, Andrew~M Dai, Anja Hauth, Katie Millican, et~al.
\newblock {Gemini: A Family of Highly Capable Multimodal Models}.
\newblock \emph{arXiv preprint arXiv:2312.11805}, 2023.

\bibitem[Team et~al.(2024)Team, Riviere, Pathak, Sessa, Hardin, Bhupatiraju, Hussenot, Mesnard, Shahriari, Ram{\'e}, et~al.]{team2024gemma2}
Gemma Team, Morgane Riviere, Shreya Pathak, Pier~Giuseppe Sessa, Cassidy Hardin, Surya Bhupatiraju, L{\'e}onard Hussenot, Thomas Mesnard, Bobak Shahriari, Alexandre Ram{\'e}, et~al.
\newblock {Gemma 2: Improving open language models at a practical size}.
\newblock \emph{arXiv preprint arXiv:2408.00118}, 2024.

\bibitem[{Textstat Contributors}(2024)]{textstat}
{Textstat Contributors}.
\newblock {Textstat: An Easy to Use Library to Calculate Statistics from Text}, 2024.
\newblock Python package for calculating readability statistics.

\bibitem[Thapliyal et~al.(2022)Thapliyal, Pont-Tuset, Chen, and Soricut]{thapliyal2022crossmodal}
Ashish~V Thapliyal, Jordi Pont-Tuset, Xi Chen, and Radu Soricut.
\newblock Crossmodal-3600: A massively multilingual multimodal evaluation dataset.
\newblock \emph{arXiv preprint arXiv:2205.12522}, 2022.

\bibitem[{\"U}st{\"u}n et~al.(2024){\"U}st{\"u}n, Aryabumi, Yong, Ko, D'souza, Onilude, Bhandari, Singh, Ooi, Kayid, et~al.]{ustun2024aya}
Ahmet {\"U}st{\"u}n, Viraat Aryabumi, Zheng-Xin Yong, Wei-Yin Ko, Daniel D'souza, Gbemileke Onilude, Neel Bhandari, Shivalika Singh, Hui-Lee Ooi, Amr Kayid, et~al.
\newblock {Aya model: An instruction finetuned open-access multilingual language model}.
\newblock \emph{arXiv preprint arXiv:2402.07827}, 2024.

\bibitem[Wang et~al.(2024)Wang, Bai, Tan, Wang, Fan, Bai, Chen, Liu, Wang, Ge, et~al.]{wang2024qwen2}
Peng Wang, Shuai Bai, Sinan Tan, Shijie Wang, Zhihao Fan, Jinze Bai, Keqin Chen, Xuejing Liu, Jialin Wang, Wenbin Ge, et~al.
\newblock {Qwen2-VL: Enhancing Vision-Language Model's Perception of the World at Any Resolution}.
\newblock \emph{arXiv preprint arXiv:2409.12191}, 2024.

\bibitem[Xiao et~al.(2024)Xiao, Wu, Xu, Dai, Hu, Lu, Zeng, Liu, and Yuan]{xiao2024florence}
Bin Xiao, Haiping Wu, Weijian Xu, Xiyang Dai, Houdong Hu, Yumao Lu, Michael Zeng, Ce Liu, and Lu Yuan.
\newblock Florence-2: Advancing a unified representation for a variety of vision tasks.
\newblock In \emph{Proceedings of the IEEE/CVF Conference on Computer Vision and Pattern Recognition}, pages 4818--4829, 2024.

\bibitem[Xue et~al.(2021)Xue, Constant, Roberts, Kale, Al-Rfou, Siddhant, Barua, and Raffel]{xue2020mt5}
Linting Xue, Noah Constant, Adam Roberts, Mihir Kale, Rami Al-Rfou, Aditya Siddhant, Aditya Barua, and Colin Raffel.
\newblock {mT5: A Massively Multilingual Pre-trained Text-to-Text Transformer}.
\newblock In \emph{Proceedings of the 2021 Conference of the North American Chapter of the Association for Computational Linguistics: Human Language Technologies}, pages 483--498, Online, 2021. Association for Computational Linguistics.

\bibitem[Young et~al.(2014)Young, Lai, Hodosh, and Hockenmaier]{young2014image}
Peter Young, Alice Lai, Micah Hodosh, and Julia Hockenmaier.
\newblock {From image descriptions to visual denotations: New similarity metrics for semantic inference over event descriptions}.
\newblock \emph{Transactions of the Association for Computational Linguistics}, 2:\penalty0 67--78, 2014.

\bibitem[Yue et~al.(2024)Yue, Song, Asai, Kim, Nyandwi, Khanuja, Kantharuban, Sutawika, Ramamoorthy, and Neubig]{yue2024pangea}
Xiang Yue, Yueqi Song, Akari Asai, Seungone Kim, Jean de~Dieu Nyandwi, Simran Khanuja, Anjali Kantharuban, Lintang Sutawika, Sathyanarayanan Ramamoorthy, and Graham Neubig.
\newblock {Pangea: A Fully Open Multilingual Multimodal LLM for 39 Languages}.
\newblock \emph{arXiv preprint arXiv:2410.16153}, 2024.

\bibitem[Zhai et~al.(2023)Zhai, Mustafa, Kolesnikov, and Beyer]{zhai2023sigmoid}
Xiaohua Zhai, Basil Mustafa, Alexander Kolesnikov, and Lucas Beyer.
\newblock {Sigmoid Loss for Language Image Pre-Training}.
\newblock In \emph{Proceedings of the IEEE/CVF International Conference on Computer Vision}, pages 11975--11986, 2023.

\bibitem[Zhang et~al.(2023)Zhang, Aljunied, Gao, Chia, and Bing]{zhang2023m3exam}
Wenxuan Zhang, Mahani Aljunied, Chang Gao, Yew~Ken Chia, and Lidong Bing.
\newblock {M3Exam: A Multilingual, Multimodal, Multilevel Benchmark for Examining Large Language Models}.
\newblock \emph{Advances in Neural Information Processing Systems}, 36:\penalty0 5484--5505, 2023.

\end{thebibliography}
}

\clearpage
\setcounter{page}{1}
\maketitlesupplementary

\section{Additional Results}
\label{sec:additional_results}
We compare Maya's responses with those of LLaVA-7B and GPT4 using a set of additional images. The objective is to qualitatively evaluate Maya's performance against a model of similar scale (LLaVA-7B) and a significantly larger model (GPT4, which is estimated to have trillions of parameters).The comparison includes examples of visual question answering and detailed caption generation to illustrate the strengths and differences among the models.

\begin{figure*}[h]
   \centering
   \begin{conversationbox}{Subway}
       \includegraphics[width=1\linewidth]{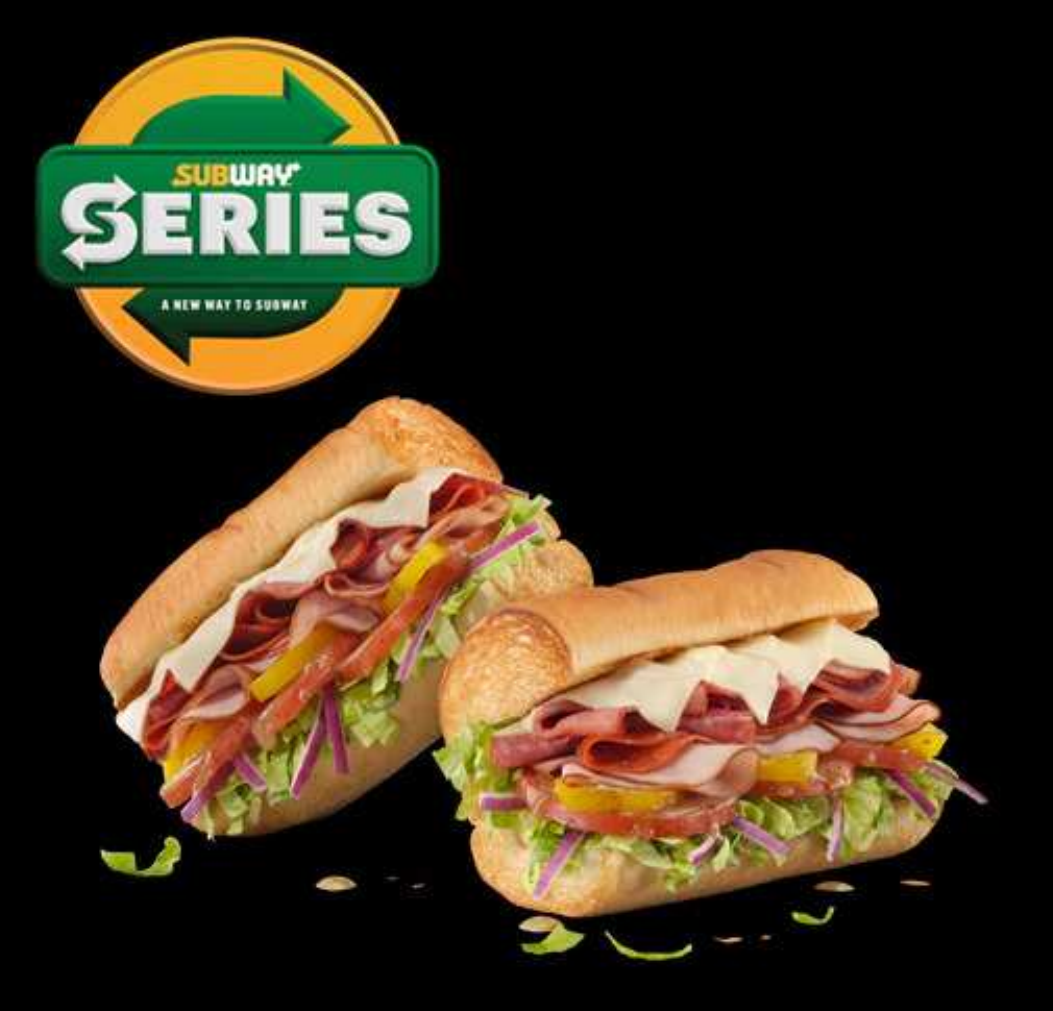}
        
       \textbf{Question:} \\
       What brand is featured in this advertisement? 
       \tcblower
       \textbf{Maya:} \\
       The Subway brand is featured in this advertisement.\\
       \textbf{LLaVA-7B:} \\
       The brand featured in this advertisement is Subway.\\
       \textbf{GPT4:} \\
       The brand featured in this advertisement is Subway.
   \end{conversationbox}
   \caption{Comparison of model responses to visual questions on brand recognition.}
   \label{fig:model_responses023}
\end{figure*}

\noindent{\textbf{Visual Question Answering.}} Figure~\ref{fig:model_responses023} presents a comparative analysis of model responses to a visual question identifying the brand featured in an advertisement. The conversation box displays the outputs from three models: Maya, LLaVA-7B, and GPT4. All three models successfully identify the brand as ``Subway". Maya demonstrates a consistent understanding of visual content and excels at extracting OCR information, particularly when the text is clearly visible and occupies a significant portion of the image.

\begin{figure*}[h]
   \centering
   \begin{conversationbox}{Diamond Head in Hawaii}
       \includegraphics[width=1\linewidth]{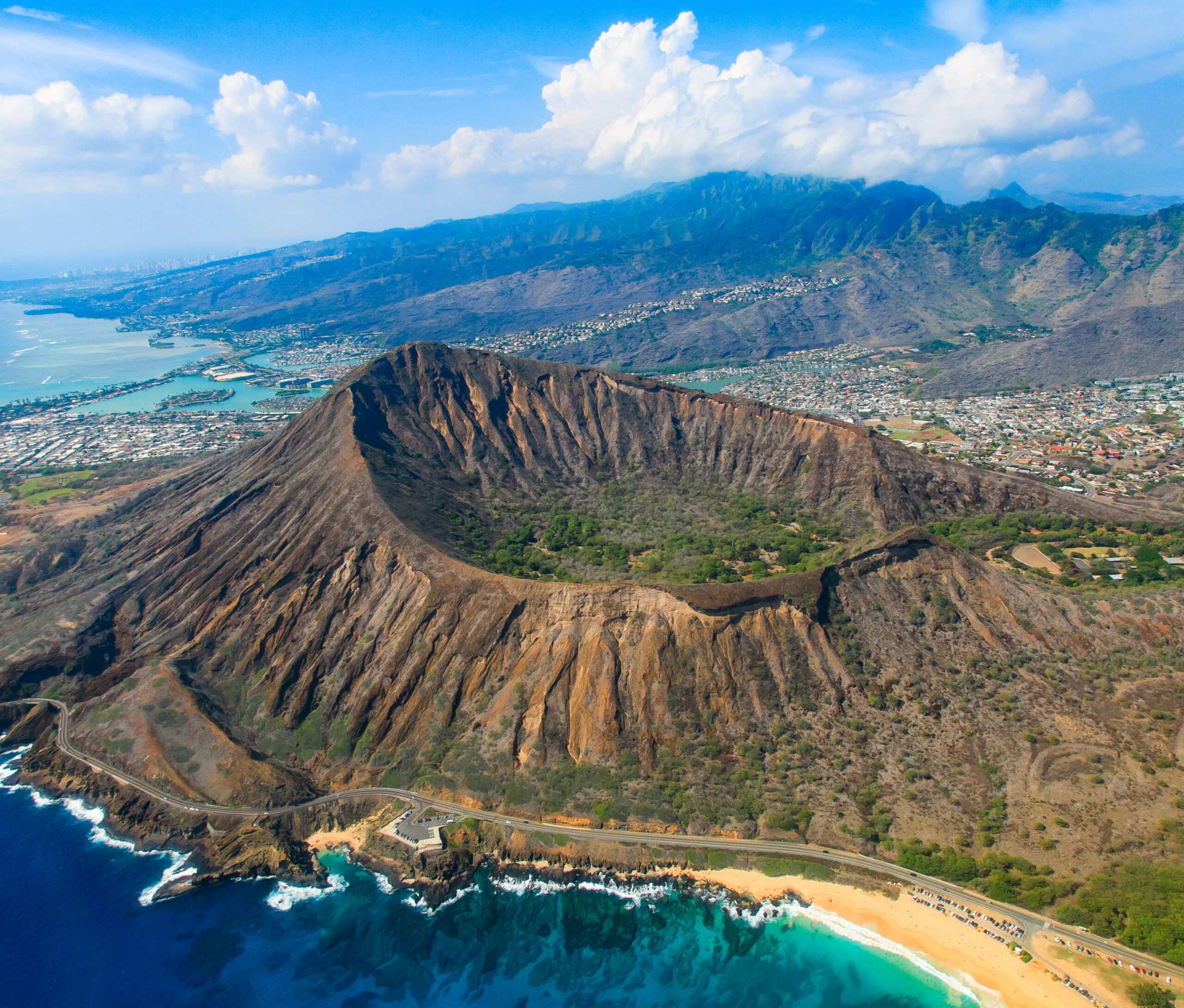}
        
       \textbf{Question:} \\
       What is the name of this famous sight in the photo? 
       \tcblower
       \textbf{Maya:} \\
       The famous sight in the photo is the Seven Islands of Hawaii.\\
       \textbf{LLaVA-7B:} \\
       The famous sight in the photo is the Diamond Head, which is a volcanic crater located in Hawaii.\\
       \textbf{GPT4:} \\
       The famous sight in the photo is Diamond Head.
   \end{conversationbox}
   \caption{Comparison of model responses to visual questions on identifying famous landmark.}
   \label{fig:model_responses001}
\end{figure*}

Figure \ref{fig:model_responses001} presents a comparative analysis of model responses in identifying a well known landmark. While LLaVA-7B and GPT4 correctly identify the landmark as ``Diamond Head", a volcanic crater in Hawaii, Maya incorrectly identifies it as the ``Seven Islands of Hawaii". This discrepancy illustrates differences in the models' accuracy and contextual understanding. The correct responses from LLaVA-7B and GPT4 demonstrate superior alignment with the visual and textual context, whereas Maya's misidentification suggests potential limitations in training on geographically specific datasets.

\begin{figure*}[h]
    \centering
    \begin{conversationbox}{Highways at Night}
        \includegraphics[width=1\linewidth]{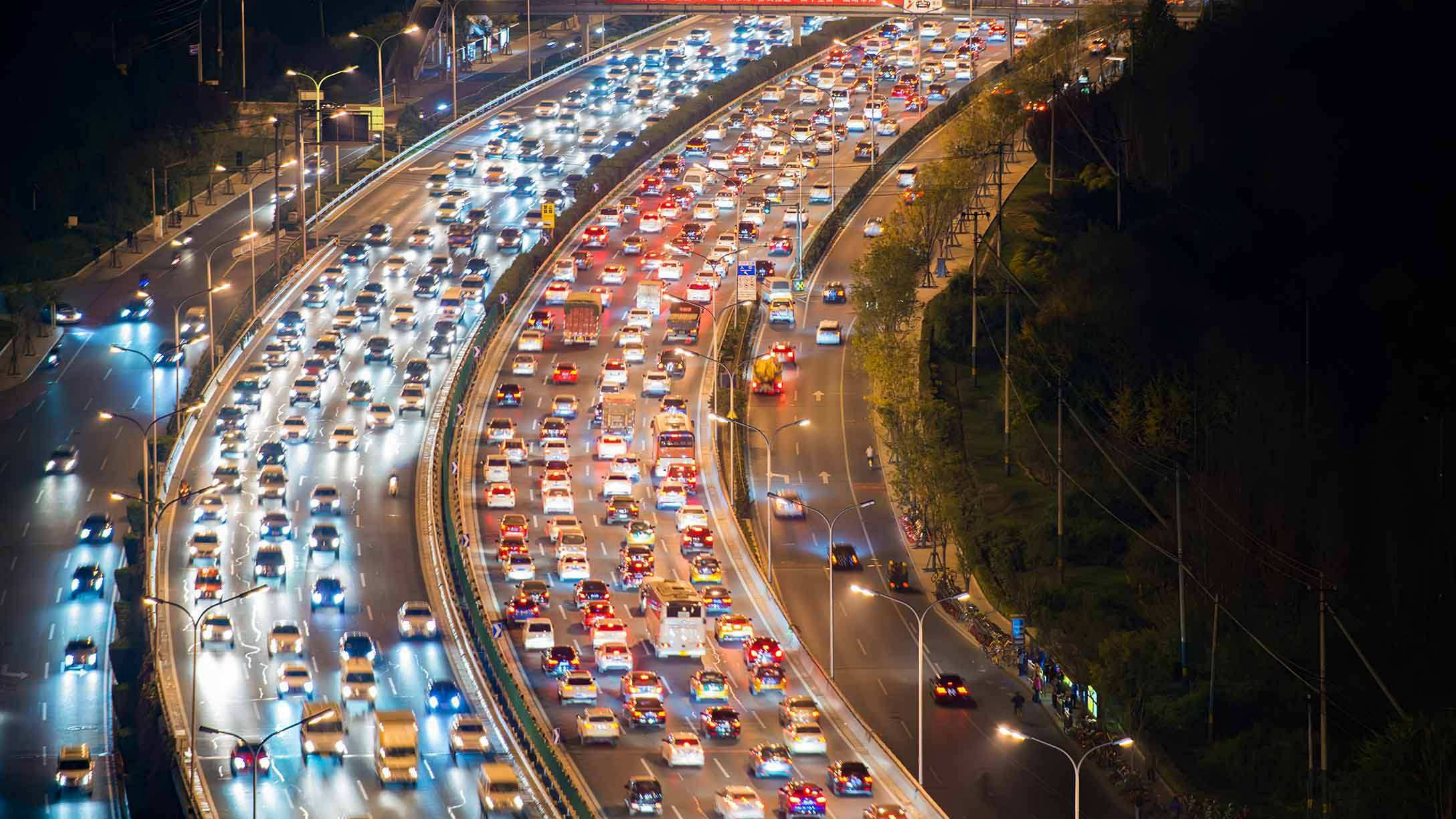}
        
        \textbf{Question:} \\
        Describe this photo in detail.
        \tcblower
        \textbf{Maya:} \\
        The image depicts a large, busy highway at night, with heavy traffic in both directions. There are numerous cars and trucks on the road, occupying almost all available lanes. The traffic appears to be moving slowly, with vehicles occupying the entire length of the highway. The scene is illuminated by the bright lights of the vehicles, as well as by the streetlights, creating a dynamic atmosphere. The highway extends into the distance, with no end in sight, highlighting the scale and congestion of this bustling transportation network.\\
        \textbf{LLaVA-7B:} \\
        The image depicts a busy highway filled with heavy traffic, including numerous cars and trucks. The highway is bustling with activity, as vehicles are seen in various positions, some moving and others waiting in line. The traffic is dense, with cars and trucks occupying almost every lane, creating a sense of congestion. The highway is surrounded by trees, adding a touch of nature to the urban landscape. The scene captures the essence of a typical day in a large city, with people commuting and traveling on the busy roads.\\
        \textbf{GPT4:} \\
        This is a top view photograph of a highway at night. The highway is divided into four sections. On the left side, there are three lanes of traffic approaching the camera and four lanes of traffic moving away from the camera. On the right side, there are four lanes of traffic approaching the camera and three lanes of traffic moving away from the camera. Most cars in the four-lane section moving away from the camera have their brake lights on, indicating a higher traffic density in the four-lane sections. Meanwhile, traffic appears to be lighter on the two three-lane sections. The four-lane highway is elevated compared to the three-lane highway, creating a visually distinct separation between the two sections. The lights alongside the highway are illuminated, providing a clear view of the road and cars. On the right side of the highway, there are trees, adding a touch of nature to the scene.
    \end{conversationbox}
    \caption{Comparison of model responses to visual questions on night highway traffic.}
    \label{fig:model_responses024_01}
\end{figure*}

\noindent{\textbf{Caption Generation.}}
Our example in Figure \ref{fig:model_responses024_01} illustrates how Maya generates a description to a highways-at-night scene. All models provide detailed description of the photo with different focuses. GPT4 provides precise information about the photo, including the number of lanes, the direction of the traffic, the density of the cars, the elevation of the four-lane highway, the illumination of the lights, and the presence of trees. Maya (similar to LLaVA-7B), while does not identify the number of lines, provides a vivid but detailed description of the scene.  

\begin{figure*}[h]
    \centering
    \begin{conversationbox}{Asian Food}
        \includegraphics[width=0.7\linewidth]{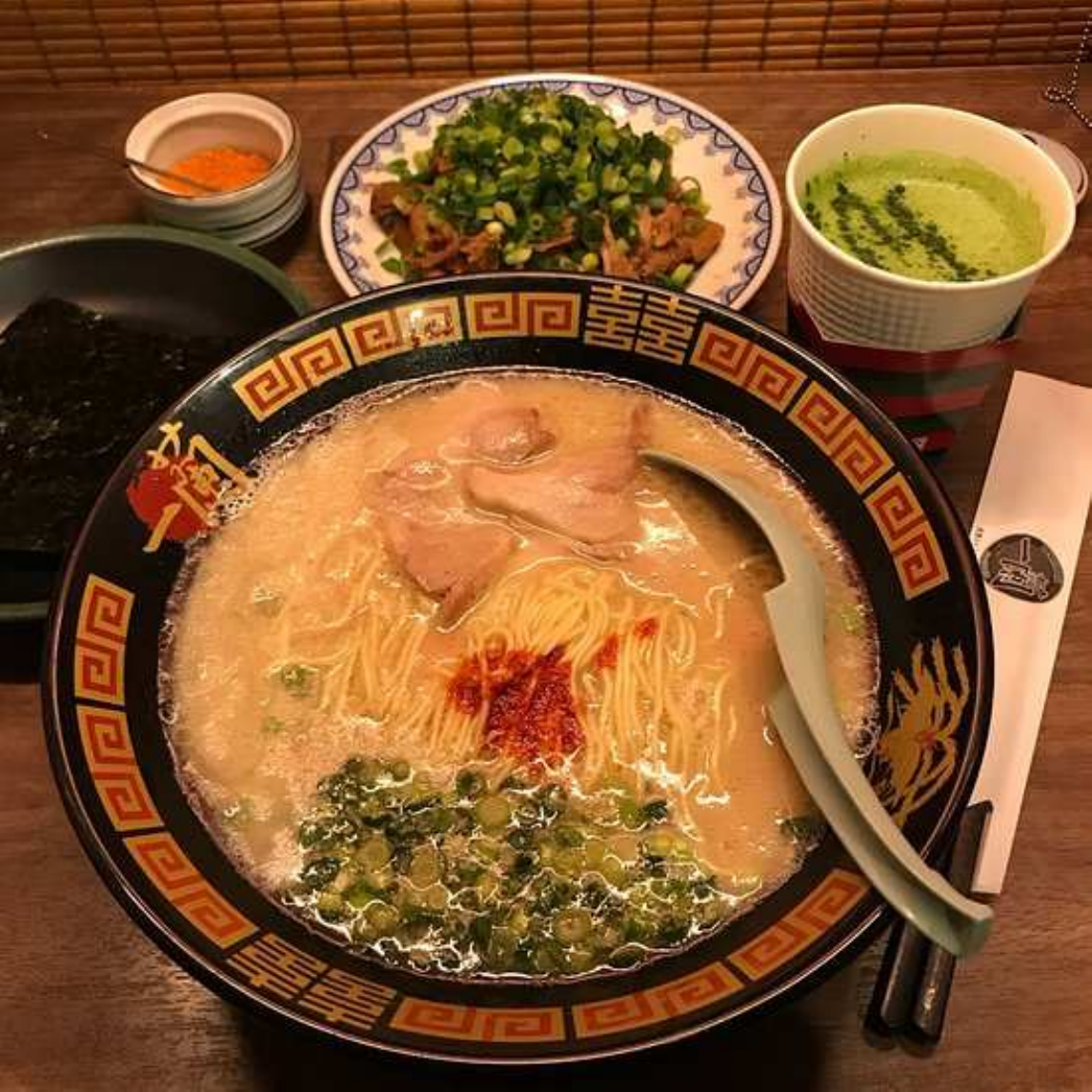}
        
        \textbf{Question:} \\
        Describe this photo in detail.
        \tcblower
        \textbf{Maya:} \\
        The image features a dining table with several bowls of food, including a large bowl of noodles and a few smaller bowls containing a variety of dishes, possibly Asian cuisine. The noodles are served in a bowl with vegetables, making it a delicious and healthy meal. There are also a couple of cups placed on the table, one near the top left corner and the other closer to the center. A spoon can be seen resting in one of the bowls, ready to be used. The table is set up with a variety of food options, making it an inviting scene for those who enjoy sharing and enjoying a meal together.\\
        \textbf{LLaVA-7B:} \\
        The image features a dining table with a variety of food items, including a large bowl of soup, a plate of noodles, and a bowl of vegetables. The soup is placed in the center of the table, while the noodles and vegetables are positioned around it. There are several bowls and cups scattered across the table, with some placed near the edges and others closer to the center. A spoon is also visible on the table, likely used for eating the soup. The table setting creates an inviting atmosphere for a meal.\\
        \textbf{GPT4:} \\
        This is a close-up photo of a meal at ICHIRAN. In the center, there is a chashu ramen bowl with a spoon. The ramen is seasoned with chili sauce and chopped scallions, and served with two pieces of chashu. To the right of the bowl, chopsticks are placed, still in their paper wrap, unopened. The ramen is accompanied by nori on the left side. On top, there are accompanying sides. From left to right, these include a bowl of orange spice (possibly garlic sauce), a plate of smoke-flavored stewed pork garnished with chopped scallions, and a cup of matcha green tea.
    \end{conversationbox}
    \caption{Comparison of model responses to visual questions on food details.}
    \label{fig:model_responses022}
\end{figure*}

Next, we focus on models' ability to describe a culturally diverse image in Figure \ref{fig:model_responses022}. Here, GPT4 accurately identify the restaurant that serves this ramen, with detailed in the photo such as chashu in the ramen, the chopsticks in their paper wrap, the garnish, and the sides served with the ramen. It shows an understanding of the context. While Maya correctly identifies different elements on the table, the description is unfortunately lacks the specific culturally relevant details provided in the context. 

\begin{figure*}[t]
  \centering
  \fbox{\includegraphics[width=1\linewidth]{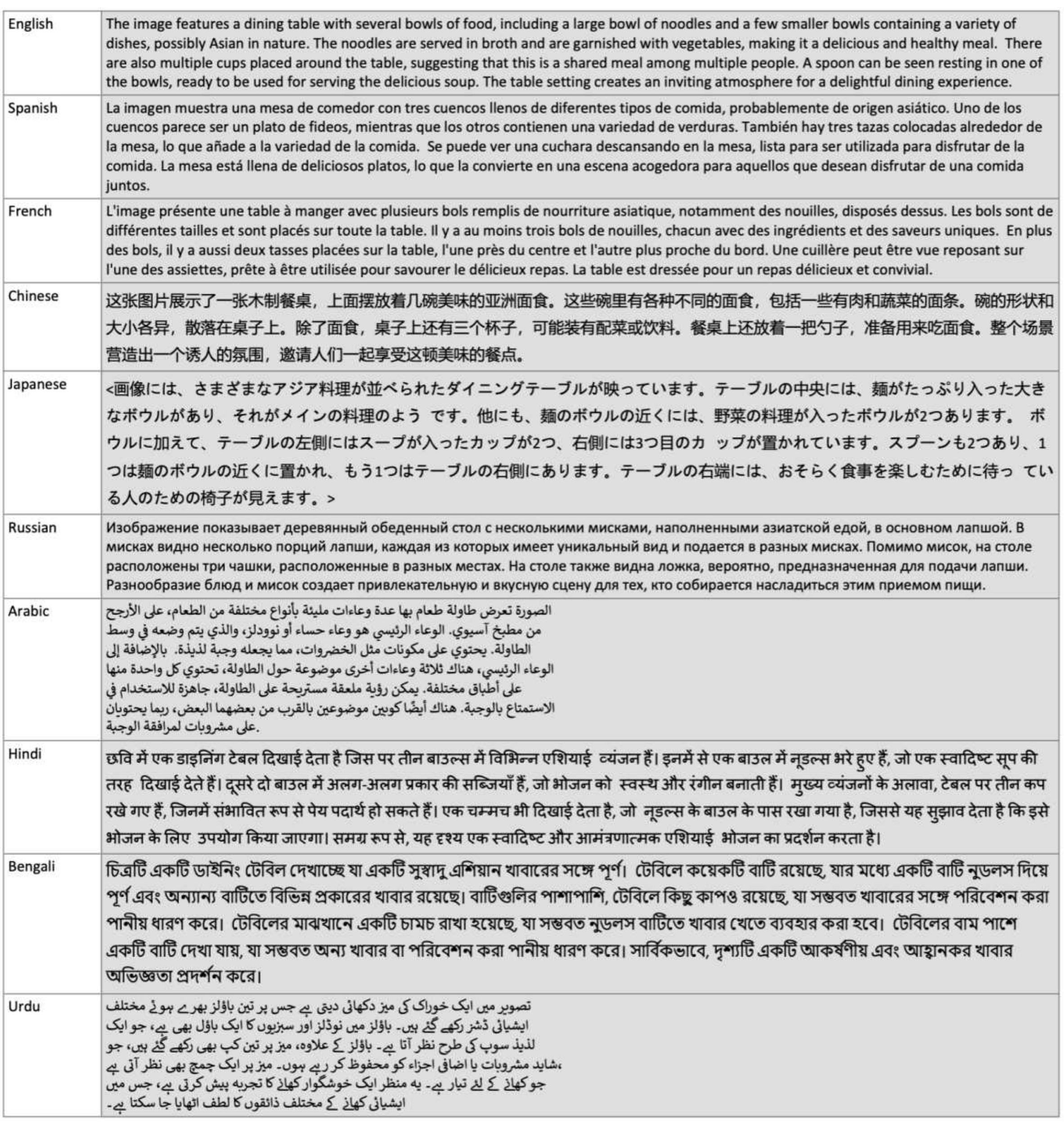}}
  \caption{Maya output for prompt (with image from Figure \ref{fig:asianfood}): Describe this photo in detail in \{language\}.}
  \label{fig:food_in_details}
\end{figure*}

\noindent{\textbf{Caption Generation in Multilingual.}}
In Figure \ref{fig:food_in_details}, we compare Maya's response in 10 languages in describing the image in Figure \ref{fig:model_responses022} in detail. Translating Hindi response to English: \textit{The image shows a dining table with three bowls containing different Asian dishes. One of the bowls has noodles, which looks like a delicious soup. The other two bowls contain different types of  vegetables, making the meal healthy and vibrant. Apart from the main dishes, there are three cups on the table, probably for beverages.  A spoon is also visible, placed near the bowl of noodles, suggesting it will be used for serving the food. Overall, this scene captures a delectable and welcoming Asian dining experience}. In Chinese, Maya describes the scene as featuring \textit{a few bowls of Asian noodles of varying kinds}, along with identifying meat, vegetables, and a spoon. It also notes the presence of \textit{three cups possibly containing side dishes or drinks}, adding that \textit{the whole scene creates an inviting atmosphere for people to enjoy this delicious meal together}. English translation of Spanish response: \textit{The image shows a dining table with three bowls filled with different types of food, probably of Asian origin. One of the bowls appears to be a bowl of noodles, while the others contain a variety of vegetables. There are also three cups placed around the table, which adds to the variety of the meal.  A spoon can be seen resting on the table, ready to be used to enjoy the meal. The table is filled with delicious dishes, making it a welcoming scene for those who want to enjoy a meal together}. Bengali response is similar to Spanish - they both fail to identify matcha, meat and does not identify the correct position of the spoon. 

%
%
%

\end{document}